\documentclass[a4paper,11pt]{article}

\usepackage{graphicx}  
\usepackage{url}       
\usepackage{times}
\usepackage[margin=25mm]{geometry}
\usepackage{color,soul} 
\usepackage{bm,amsmath,amssymb} 
\usepackage{psfrag}

\def\internalversion{0}

\newcommand{\ie}{\mbox{\emph{i.e.\ }}}
\newcommand{\wrt}{\mbox{\emph{w.r.t.\ }}}
\newcommand{\eg}{\mbox{\emph{e.g.\ }}}
\newcommand{\etc}{\mbox{\emph{etc.}}}

\newcommand{\Eq}[1]{Eq.\,(\ref{#1})}  
\newcommand{\Fig}[1]{Fig.\,\ref{#1}}  
\newcommand{\Sec}[1]{Sec.\,\ref{#1}}  
\newcommand{\Tab}[1]{Tab.\,\ref{#1}}  

\newcommand{\M}[1]{\mathtt{#1}}   
\newcommand{\V}[1]{\mathbf{#1}}   
\newcommand{\C}[1]{\mathcal{#1}}  

\newcommand{\imu}{\textsc{imu}}  

\newcommand{\MM}[1]{\textsc{#1}}  

\newcommand{\rul}[1]{\textcolor{red}{\underline{\textcolor{black}{#1}}}}

\newcommand{\Figs}{Figs/}

\title{Ego-Motion Alignment from Face Detections\\ for Collaborative Augmented Reality}
\date{}

\author{Branislav Micusik \quad \quad \quad Georgios Evengelidis\\
       Snap Inc.\\
       Vienna, Austria \\
       \texttt{\small brano@snap.com \quad  georgios@snap.com}
}

\begin{document}
\maketitle

\begin{abstract}%
Sharing virtual content among multiple smart glasses wearers is an essential feature of a seamless Collaborative Augmented Reality experience. To enable the sharing, local coordinate systems of the underlying 6\MM{d} ego-pose trackers, running independently on each set of glasses, have to be spatially and temporally aligned with respect to each other. In this paper, we propose a novel lightweight solution for this problem, which is referred as ego-motion alignment. We show that detecting each other's face or glasses together with tracker ego-poses sufficiently conditions the problem to spatially relate local coordinate systems. Importantly, the detected glasses can serve as reliable anchors to bring sufficient accuracy for the targeted practical use. The proposed idea allows us to abandon the traditional visual localization step with fiducial markers or scene points as anchors. A novel closed form minimal solver which solves a Quadratic Eigenvalue Problem is derived and its refinement with Gaussian Belief Propagation is introduced. Experiments validate the presented approach and show its high practical potential.  
\end{abstract}

\section{Introduction}
\begin{figure}[t]
	\begin{center}
		\scriptsize
		\psfrag{AA}[l][l]{$\C{O^A}$}
		\psfrag{BB}[r][r]{$\C{O^B}$}
		\begin{tabular}{c@{\hspace{10mm}}c}
		    \psfrag{scene}[tc][tc]{\MM{scene}}
			\includegraphics[width=0.45\linewidth]{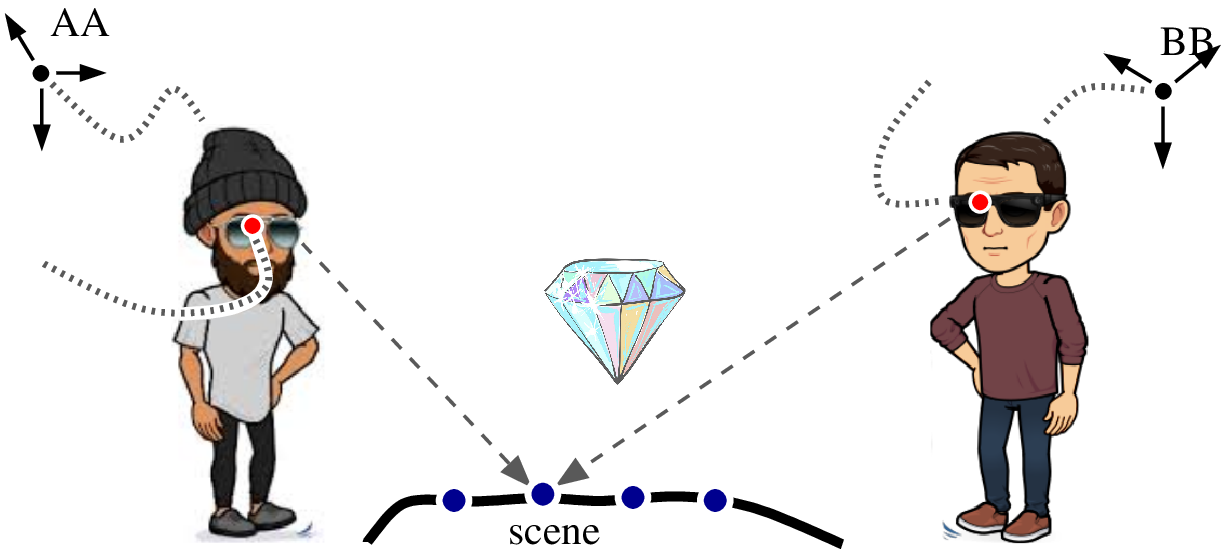} &
			\includegraphics[width=0.45\linewidth]{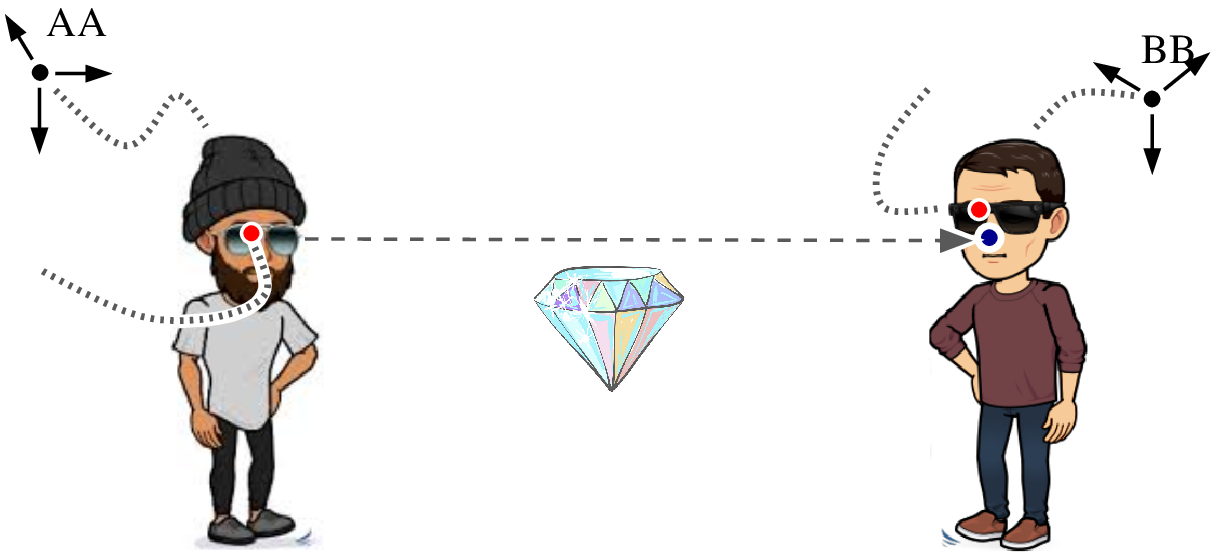} \\[2ex]
			\normalsize (a) \MM{Standard} & \normalsize (b) \MM{Proposed}\\[-4ex]
		\end{tabular}
	\end{center}
	\caption{Standard vs. proposed principle to align local coordinate systems of AR glasses wearers for Collaborative AR. The alignment allows the wearers to see a virtual object, \eg a diamond, at the same scene location. (a) Standard solutions use points of the surrounding rigid scene as anchors. (b) The proposed solution uses each other's facial landmarks as anchors instead.}
	\label{fig:motivation_comparison}
\end{figure}

\begin{figure}[t]
  \begin{center}
  	 \psfrag{A}[r][r]{\large \MM{a}}
  	 \psfrag{B}[br][Br]{\large \MM{b}}
     \scriptsize
     \psfrag{AA}[l][l]{$\C{O^A}$}
     \psfrag{BB}[l][l]{$\C{O^B}$}
     \psfrag{R, t}[t][t]{$?\ \M{R}_\C{B}^\C{A}, \V{t}_\C{B}^\C{A}$}
     \includegraphics[width=0.8\linewidth]{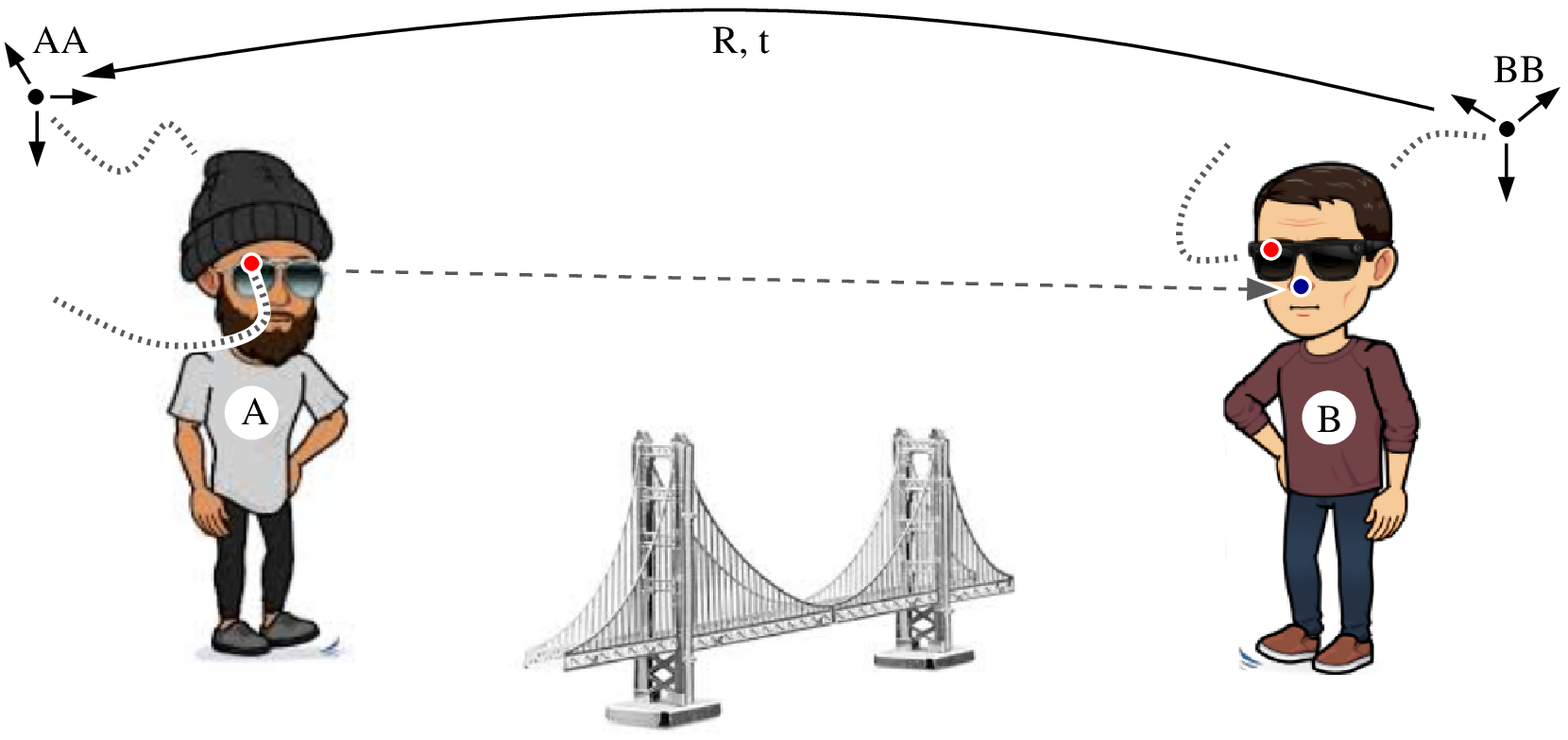} \\
     \begin{tabular}{c@{\hspace{5cm}}c}
      \psfrag{camA}[lb][lb]{\MM{cam} A}
      \includegraphics[width=0.18\linewidth]{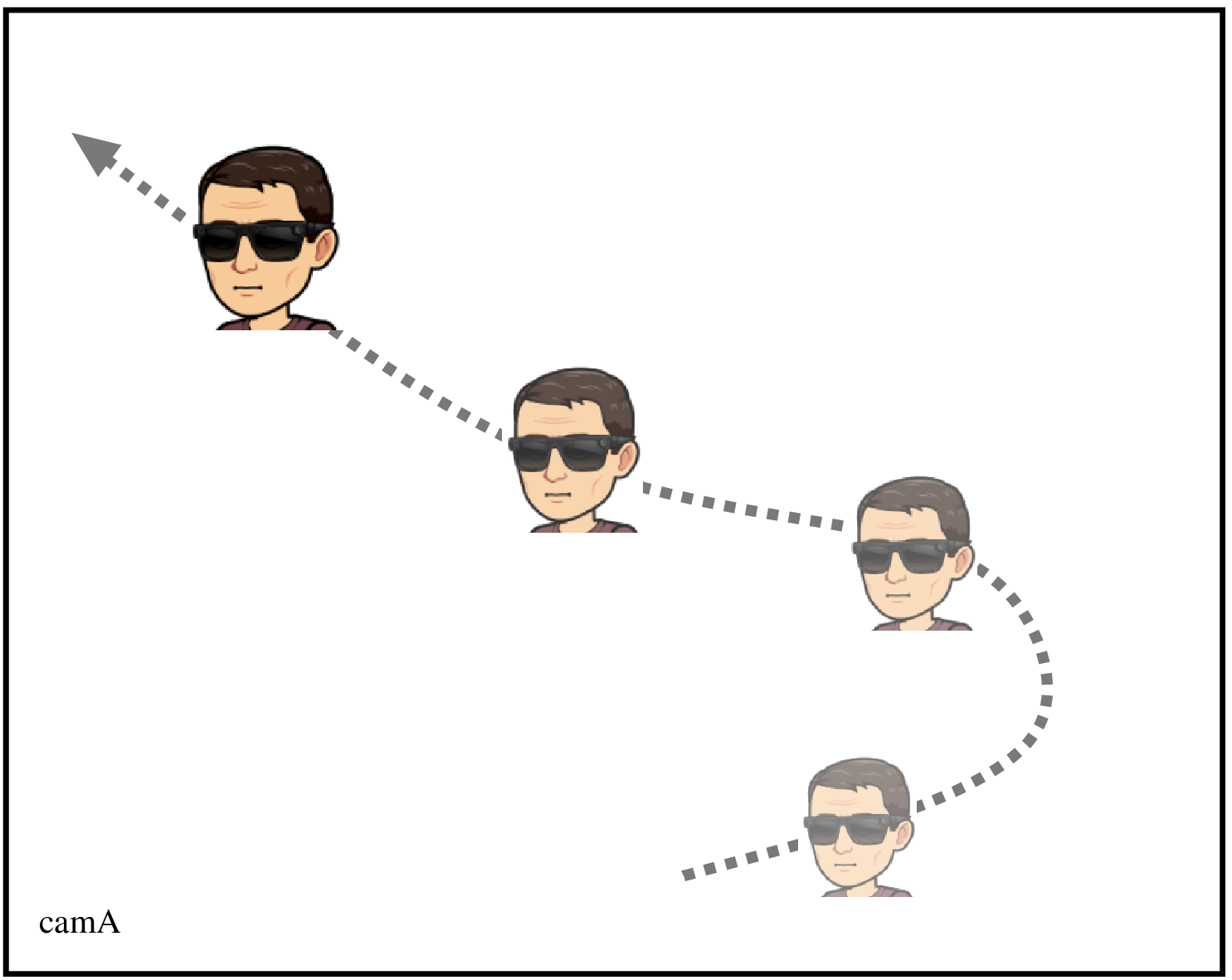} &
      \psfrag{camB}[lb][lb]{\MM{cam} B}
      \includegraphics[width=0.18\linewidth]{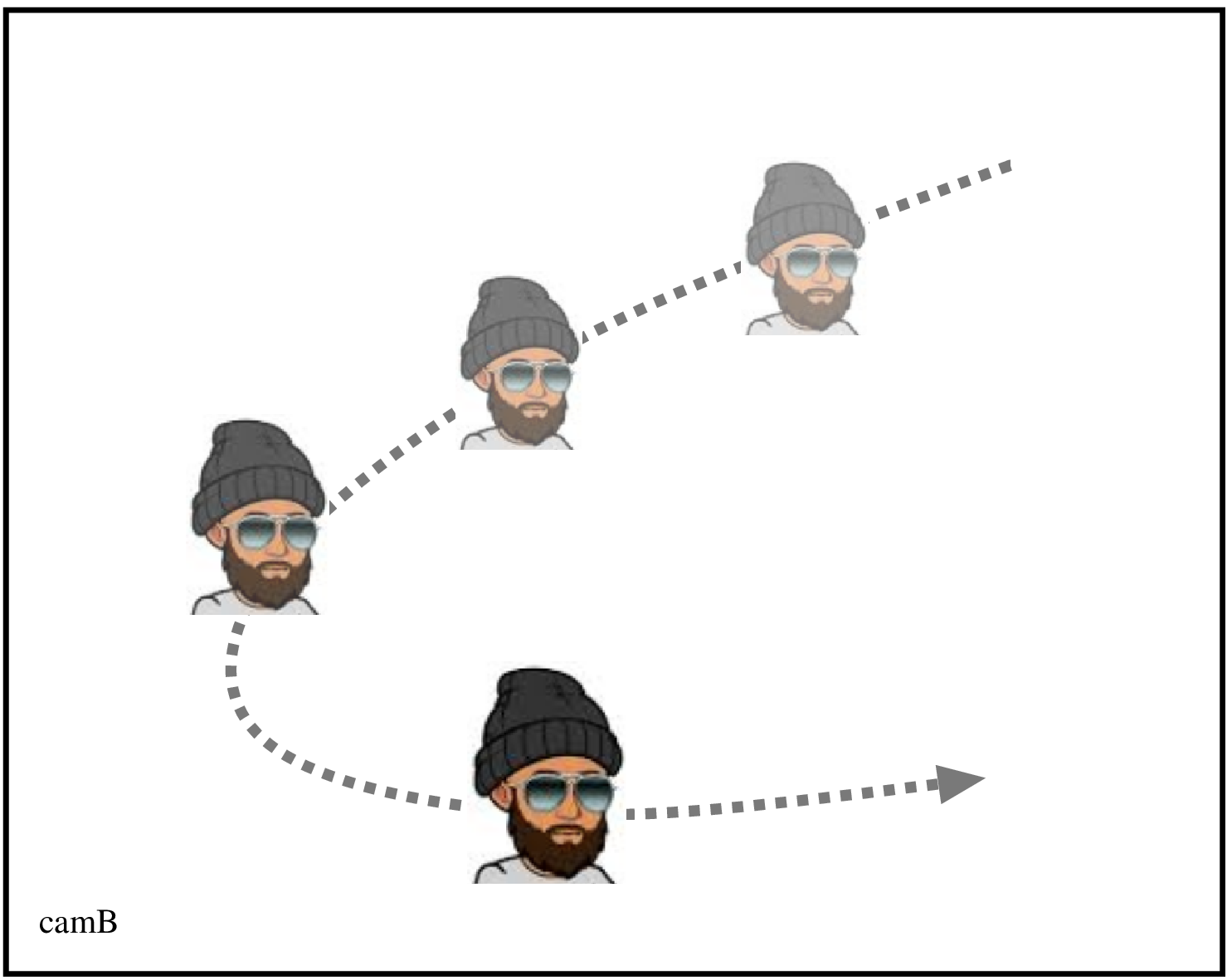} 
     \end{tabular}
  \end{center}
  \caption{Collaborative AR: two AR glasses wearers look at and interact with the same virtual 3\MM{d} content, a bridge. In order to share AR experience between users, their local coordinate systems need to be aligned by an {\em unknown} transformation into a common coordinate system, chosen to be $\C{A}$. The 6\MM{d} ego-poses of the red points are known in each user's local coordinate system. By tracking a single or multiple facial landmarks of another user, shown in blue, the alignment of local coordinate systems is shown in this paper to be solvable. The relation of the blue tracked point \wrt the body coordinate system's origin shown as the red dot is rigid but can be unknown.}
  \label{fig:motivation}
\end{figure}

Augmented Reality (AR) has been becoming ubiquitous and silently penetrating into our daily lives. It has been boosted by the growth of computing power on mobile phones and by progress on miniaturization of the form factor of wearable smart glasses. AR is one of the use cases of the Spatial AI effort, along with autonomous driving. Spatial AI attracts a lot of attention in the scientific community and has seen tremendous achievements with the advent of deep learning. This pushes novel features and functionalities into AR systems. Specifically, AR is the  mixing of the real physical world with the virtual one indistinguishably, such that the canvas becomes the entire world. A unique feature of AR is that people share scene augmentations and can collaborate together. Collaborative AR allows seamless sharing of the same 3\MM{d} content between multiple AR users. Any user can augment a shared virtual space with new content which others immediately see and may change. A requirement for collaborative AR to react appropriately is that the system knows at each time instance the relative position and orientation of the users.

AR devices retrieve their ego-motion by running a 6\MM{d} ego-pose tracker (VIO, SLAM, SAM, \etc) as a local background service. The ego-pose tracker gives, at a certain temporal cadence, the translation and orientation of the body coordinate system, typically \imu{}'s, in the local world. Ego-pose trackers define origins of their local coordinate systems at the time and place of the start of operation. Collaborative AR  requires that these local coordinate systems are aligned such that each ego-pose in the local world can be transformed to a common global coordinate system. This problem is called the  Ego-Motion Alignment Problem which is tackled in this paper.

A standard approach to solve the alignment problem is to use fiducial markers or scene 3\MM{d} points, see \Fig{fig:motivation_comparison}(a). In the latter, the reconstructed local world around the user as a 3\MM{d} point cloud is shared such that the other users can localize in that world. Running a full mapping service along with the ego-pose tracker is an expensive process due to computational burden and large memory footprint for mobile devices with limited battery life. Moreover, privacy issues may arise when sharing the point cloud with image point descriptors. From the algorithmic point of view, building such a model and then re-localizing in it has not been fully solved yet. Success of mapping and re-localization strongly depends on the viewing angle and scene properties like textureness, repetitions, and rigidity. Moreover, a very common situation in social interaction is when users stand in front of each other. The overlapping scene area in this case, which could be used for standard visual localization, is very small and is observed form very different vantage points. This makes the standard localization methods prone to fail. We therefore propose a different solution, well suited for such a use case and for collaborative AR on smart glasses.

AR on wearable smart glasses offer a unique feature as opposed to AR on smart phones. Glasses keep the device fixed at the wearer's head during their operation. This fact allows to use human faces or glasses as anchors to register multiple devices in one common coordinate system. Let us assume two persons wearing smart glasses and looking at the same part of the environment while shortly observing each other, see \Fig{fig:motivation_comparison}(b). We propose to align their local coordinate systems by tracking each other's face and to employ this information into a newly designed solver. The proposed method brings a lightweight solution to the alignment problem as it only requires to share the tracklets of a point on the face or glasses and 6\MM{d} ego-poses at these locations, as shown in \Fig{fig:motivation}. 

The main difference of the proposed solution to standard techniques is that the anchors are faces or glasses and not the $3$\MM{d} scene itself, shown in \Fig{fig:motivation_comparison}. The scene is utilized only for the $6$\MM{d} ego-pose tracker, but independently for each device and with possibly non-overlapping scene parts. Our contribution is threefold. 
\begin{itemize}
	\item We show for the first time that tracklets of a point on the face symmetry plane or glasses of another user and local $6$\MM{d} ego-poses give sufficient constraints to solve the ego-motion alignment problem. 
	\item We propose a novel closed form minimal solver based on solving a Quadratic Eigenvalue Problem. 
	\item We cast refinement of the initial closed form solution into the probabilistic framework as Gaussian Belief Propagation inference in a Factor Graph. This inference strategy is very well suited for massive parallel processing architectures. 
\end{itemize}

The paper is organized as follows. We introduce the ego-motion alignment problem in \Sec{sec:ego_motion_alignment} and in its subsections we derive a closed-form solution for the two user case, with proof on feasible solutions, and necessary constraints towards solvability. For practical reasons of sufficient robustness against noise in face and glasses detections, a bi-directional constraint is added in \Sec{sec:bidir}. We present a statistically optimal refinement stage as a probabilistic framework in \Sec{sec:refinement}. Face and glasses tracker is discribed in \Sec{sec:faceGlassesTracker}. Experiments finally demonstrate great practicality of the proposed method in \Sec{sec:experiments}.
\section{Related Work}

Standard approaches formulate the ego-motion alignment as the visual localization problem. The basic principle is to use a 3\MM{d} scene as an anchor element to calculate the relative transformations. First, a point cloud of a scene is reconstructed. Second, a query camera is localized in that point cloud by matching its descriptors to the descriptors of the query camera image. There are traditional, old-school approaches which tackle the problem by explicit geometric modeling with image descriptors, \eg \cite{Cao-CVPR2014}, or without descriptors, \eg \cite{Micusik-CVPR2015,Liu-ICCV2017}. Another group of approaches are newer, deep learning based, methods of various forms. Some regress directly pose in  end-to-end fashion just from the input images, \eg a pioneering work of \cite{Kendall-ICCV2015}, some learn descriptors together with re-localization, \eg \cite{Du-ECCV2020}, some combine learning with explicit geometric modelling, \eg \cite{Brachmann-CVPR2018}. A nice comparison of both approaches can be found in \cite{Sattler-CVPR2019}. A broad overview of the entire huge visual localization topic is well reviewed in \cite{Piasco-PR2018}.

In addition, visual localization for wearable devices poses a privacy issue, as sharing whole images between users might not be allowed. Sharing just image descriptors of 3\MM{d} points has been shown to be sufficient to reconstruct the point cloud~\cite{Pittaluga-CVPR2018}, and might not be allowed as well. To facilitate this problem, a privacy preserving line reconstruction is proposed in \cite{Speciale-CVPR2019} and further developed in \cite{Geppert-ECCV2020}.

In summary, visual localization cannot fully solve the problem as it brings many challenges when considering computational burden, memory footprint, battery consumption, and robustness. Our approach therefore proposes to abandon the standard scene anchors and use user faces or glasses instead. A moving person has been used as an anchor, and so as the calibration target, to relate spatially multiple cameras in \cite{Caspi-IJCV2002, Rahimi-CVPR2004, Micusik-CVPR2011, Micusik-ICCV2011}. In the line of work \cite{Micusik-CVPR2011, Micusik-ICCV2011} static cameras with unknown person's ego-motion are assumed. The smoothness constraint is enforced to turn the ill-posed problem into a solvable one. In the ego-motion problem which is tackled in this paper, we assume moving cameras, but with known ego-poses and no explicit smoothness constraint. The problem of~\cite{Micusik-CVPR2011} also builds on knowing gravity direction and leads to the Quadratic Eigenvalue Problem. It is interesting to see how these problems are relevant, despite assuming different inputs.

Geometrically, the tackled ego-motion alignment problem is similar to the relative pose problem of generalized cameras in
\cite{Sweeney-3DV2014}. However, in the ego-motion alignment of this paper, the tracked point moves and we do not know the relation of that point to the body origin with known 6\MM{d} ego-poses. The minimal solver of \cite{Sweeney-3DV2014} also uses the gravity vector as a practical solution to turn the full non-linear problem solvable. They relax the general Gr\"obner Basis based solution into a simplified, more robust, one which is cast as an Eigenvalue Problem. Enforcing gravity direction in pose estimation problems is a very practical constraint for many use cases. It was proposed by \cite{Kukelova-ACCV2010} for camera absolute pose and followed by \cite{Micusik-CVPR2011,Lee-CVPR2014, Sweeney-3DV2014} for relative camera pose problems.

Maximum likelihood refinement of a closed form solution, or simply Bundle Adjustment, is a classic geometric problem in Photogrammetry~\cite{Foerstner-Book2016} and Computer Vision~\cite{Hartley-Book2004,Lourakis-TMS2009}. It is typically solved as a non-linear gradient based optimization technique on the image re-projection error. It performs iterative re-linearization which yields a sparse linear system of normal equations. Another line of thinking to tackle optimization problems are probabilistic graphical models and inference which estimates MAP of the graph vertex state~\cite{Bishop-MLbook2006, Koller-Book2009}. It was shown in~\cite{Ranganathan-IJCAI2007} how to formulate BA with a Factor Graph for the Smoothing and Mapping (SAM) problem of SLAM. We cast the ego-motion alignment problem into such a probabilistic framework and review the most relevant papers in \Sec{sec:refinement}. 

\section{Ego-Motion Alignment}
\label{sec:ego_motion_alignment}
In this section, we formulate the ego-motion alignment problem from the geometric point of view and  accordingly derive a closed-form minimal and overconstrained solver. For the sake of simplifying the presentation and derivations, we consider two glasses wearers. Including more is a straightforward augmentation. We assume that
\begin{itemize}
	\item smart glasses possess a camera and an \imu{} sensor. The camera is calibrated internally and externally \wrt to the \imu{} sensor. 
	\item  each glasses run a visual-inertial odometry (VIO) tracker which delivers a 6\MM{d} ego-pose of the \imu{} origin in the local coordinate system, set at the start of each glasses operation. Any off-the-shelf VIO system can be used. 
	\item  each glasses run a face and optionally glasses detector and tracker on top of it in order to track a point on the opposite user's face or glasses, as \Fig{fig:motivation} depicts. Several tracklets of the same point may be available which means that the opposite glasses wearer is not visible in all frames. Wearers do not need to see themselves simultaneously. 
	\item 2\MM{d} image detections and tracklets of a face or glasses together with 6\MM{d} ego-poses of the camera which perceived them for all wearers are available at one place to be processed. It is foreseen to be one of the glasses but can also be a cloud server.
	\item the ego-pose trackers of multiple users are mutually time synchronized.
	\item the alignment procedure is foreseen to run at the beginning of a session, \eg in a couple of seconds. However, any re-calculation and continuous refinement is possible during the session. 
\end{itemize}
These assumptions can be met for most of the available smart AR glasses on the market. 

\begin{figure}[t]
	\begin{center}
		\scriptsize
		\psfrag{AA}[tc][tl]{$\C{O^A}$}
		\psfrag{BB}[tl][tl]{$\C{O^B}$}
		\psfrag{Oi}[tl][tl]{$\C{O^I}$}
		\psfrag{R, t}[r][l]{$?\ \M{R}_\C{B}^\C{A}, \V{t}_\C{B}^\C{A}$}
		\psfrag{Ri, ti}[c][c]{$\M{R}_\C{I}^\C{B}, \V{t}_\C{I}^\C{B}$}
		\psfrag{Rc, tc}[c][c]{$\M{R}_\C{A}^\C{C}, \V{t}_\C{A}^\C{C}$}
		\psfrag{pa}{$\hat{\V{p}}^\C{C}$}
		\psfrag{pb}{}
		\psfrag{L}{$\V{L}^\C{I}$}
		\psfrag{K}{$\V{K}^\C{I}$}
		\psfrag{imu}[bl][bl][1][-15]{\MM{imu}}
		\psfrag{camera}[bl][bl]{\MM{camera}}
		\includegraphics[width=0.6\linewidth]{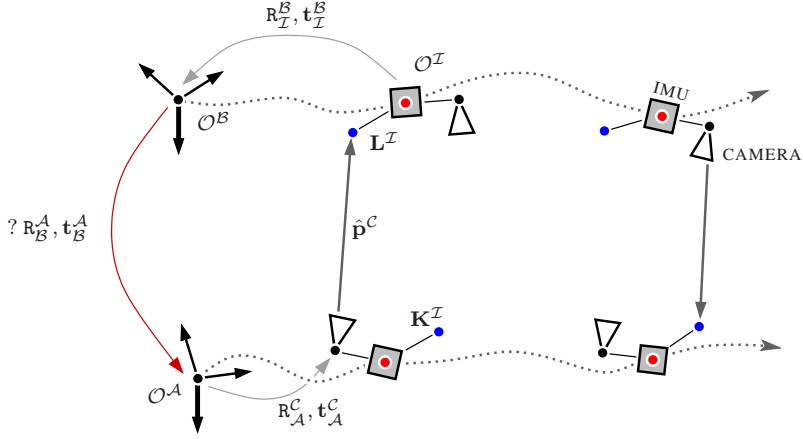} \\
	\end{center}
	\caption{Geometry of the ego-motion problem. \MM{camera}-\MM{imu} rig $\C{B}$ moves and its time evolving 6\MM{d} ego-poses $\M{R}_\C{I}^\C{B}, \V{t}_\C{I}^\C{B}$ are expressed in the local origin $\C{O^B}$, similarly for the rig $\C{A}$. The unknown relative pose $\M{R}_\C{B}^\C{A}, \V{t}_\C{B}^\C{A}$ between the local coordinate systems is unknown and subject to be estimated. }
	\label{fig:geometry}
\end{figure}

\subsection{Minimal solver}
\label{sec:minsolver}
Following \Fig{fig:geometry}, let us assume two local VIO coordinate systems $\C{A}$ and $\C{B}$ of two moving glasses, with origins $\C{O^A}$ and $\C{O^B}$ such that 
\begin{equation}
  \V{X}^\C{A} = \M{R}_\C{B}^\C{A} \V{X}^\C{B} + \V{t}_\C{B}^\C{A}
\end{equation}
transforms the same point $\V{X}$ from base $\C{B}$ to $\C{A}$. The body coordinate system is placed into the coordinate system of the \imu{}, see also \Fig{fig:head}(a). Let us assume that a 3\MM{d} point $\V{L}$ is rigidly mounted to the body frame and can be transformed from the body, \ie from the \imu{} $\C{I}$, into $\C{B}$ as
\begin{equation}
   \V{L}^\C{B} = \M{R}_\C{I}^\C{B} \V{L}^\C{I} + \V{t}_\C{I}^\C{B}.
\end{equation}
The lever arm $\V{L}^\C{I}$ is an {\em unknown} constant which can further be expressed in base $\C{A}$ as
\begin{equation}
\V{L}^\C{A} = \M{R}_\C{B}^\C{A} \V{L}^\C{B} + \V{t}_\C{B}^\C{A}.
\end{equation}
The rotation $\M{R}_\C{I}^\C{B}$ and the translation $\V{t}_\C{I}^\C{B}$ represent the known 6\MM{d} VIO ego-pose at time $t$ and should be correctly written as $\M{R}_\C{I}^\C{B}(t)$, $\V{t}_\C{I}^\C{B}(t)$. For the sake of simplicity, we drop writing the time dependency.  The point $\V{L}^\C{A}$ is expressed in the second camera as
\begin{equation}
  \lambda \hat{\V{p}}^\C{C} = \M{R}_\C{A}^\C{C} \V{L}^\C{A} + \V{t}_\C{A}^\C{C},
  \label{eq:full_projection}
\end{equation}
where $\lambda$ is a scale, $\M{R}_\C{A}^\C{C}$ and $\V{t}_\C{A}^\C{C}$ represent the pose of camera $\C{A}$ in its local coordinate system, and is known through the VIO ego-pose. More precisely, the camera bases $\C{C}$ should be indexed by the camera indices too, \ie by $\C{C_A}$ or $\C{C_B}$, and should be written $\M{R}_\C{A}^\C{C_{\!\!A}}(t)$. We drop the precise form to simplify the notation as it is evident from the context which camera is meant. Putting it all together, the point $\V{L}^\C{I}$ should be viewed by camera $\C{A}$ as
\begin{equation}
\hat{\V{p}}^\C{C} = \M{R}_\C{A}^\C{C} \left( \rul{\M{R}_\C{B}^\C{A}} \left( \M{R}_\C{I}^\C{B} \rul{\V{L}_{\ }^\C{I}} + \V{t}_\C{I}^\C{B} \right) + \rul{\V{t}_\C{B}^\C{A}} \right) + \V{t}_\C{A}^\C{C}.
\label{eq:projected_ray}
\end{equation}
The relative transformation with $\M{R}_\C{B}^\C{A}$, $\V{t}_\C{B}^\C{A}$ and the lever arm translation $\V{L}^\C{I}$ are {\em unknown} and subject to be estimated. This yields highly non-linear problem which could potentially be solved by Gr\"obner Basis. It is known that the solution of such a non-linear system of equations is very noise sensitive. Therefore, we relax the problem and  make the following practical assumption. VIO typically has its coordinate system aligned with the gravity direction as it is indirectly obtained from \imu{} measurements due to constant force towards the earth. The gravity fixes two angles, the tilt (pitch) and the roll. Therefore we can safely assume that the unknown rotation $\M{R}_\C{B}^\C{A}$ is the rotation around the gravity vector only by a pan (yaw) angle and can be parametrized by one parameter $s$. We use the following quaternion based parametrization
\begin{equation}
\M{R}_\C{B}^\C{A} = \frac{1}{1+s^2} \left( 2(\V{g}\V{g}^\top + s [\V{g}]_{_\times}) + (s^2 - 1) \M{I}_3 \right),
\label{eq:rot_parametrization}
\end{equation}
where $\V{g}$ is the unit vector representing the axis of rotation and $\M{I}_3$ a $3\times3$ identity matrix.  In our case, $\V{g}$ is the gravity vector, expressed in $\C{O_A}$. See Eq.(8.54) of F{\" o}rstner \cite{Foerstner-Book2016} for more details. Then, putting \Eq{eq:rot_parametrization} into \Eq{eq:projected_ray} gives
\begin{align}
(1+s^2) \M{R}_\C{C}^\C{A} \, \hat{\V{p}}^\C{C} = &  & \big((2 \V{g}\V{g}^\top - \M{I}) \M{R}_\C{I}^\C{B} \, & \V{L}^\C{I} & %
                                    + & \V{t}_\C{B}^\C{A} &%
                                    + (2 \V{g}\V{g}^\top - \M{I}) \, \V{t}_\C{I}^\C{B} + \M{R}_\C{C}^\C{A} \V{t}_\C{A}^\C{C}  &\big) \hspace*{2.5mm} + \notag\\
                                 & & + \big(  2\, [\V{g}]_{_\times} \M{R}_\C{I}^\C{B} \, & \V{L}^\C{I} &%
                                        & & %
				    + 2\, [\V{g}]_{_\times}  \V{t}_\C{I}^\C{B} & \big)\,s + \notag\\
                                 & & + \big( \M{R}_\C{I}^\C{B} \, & \V{L}^\C{I} & %
                                        + & \V{t}_\C{B}^\C{A} &%
                                        + \V{t}_\C{I}^\C{B} + \M{R}_\C{C}^\C{A} \, \V{t}_\C{A}^\C{C} & \big)\, s^{2}.%
   \label{eq:basic_equation}
\end{align}
The lever arm $\V{L}$ is observed in camera $\C{A}$ as $\V{u}^\C{A}$. Recall that this is the point which is being tracked in the camera image, either the point on a face or glasses. The image point $\V{u}^\C{A}$ can be brought into the calibrated spherical representation as
\begin{equation}
\V{p} = \C{N}(\M{K}_\C{A}^{-1} \V{u}^\C{A}),
\label{eq:seen_ray}
\end{equation}
where $\C{N}(\V{x})=\frac{\V{x}}{\|\V{x}\|}$ is the normalization function on the vector to its unit length and $\M{K}_\C{A}$ stands for the camera calibration matrix of camera $\C{A}$.

The vectors $(1+s^2) \M{R}_\C{C}^\C{A} \hat{\V{p}}^\C{C}$ from \Eq{eq:projected_ray} and $\lambda \V{p}$ from \Eq{eq:seen_ray} should be the same, up to the detection and calibration errors. We eliminate their scales and enforce their directions to align, so
\begin{align}
	\lambda \V{p} \times \hat{\V{p}}^\C{C}  &= 0 \notag\\
\lambda \, \M{R}_\C{C}^\C{A} \V{p} \times (1+s^2) \M{R}_\C{C}^\C{A} \hat{\V{p}}^\C{C}  &= 0 \notag\\
	[\M{R}_\C{C}^\C{A} \, \V{p}]_{_\times}\, (1+s^2) \M{R}_\C{C}^\C{A} \hat{\V{p}}^\C{C} & = 0
	\label{eq:cost}
\end{align}
This gives two linearly independent equations and so two constraints. \Eq{eq:cost} and \Eq{eq:basic_equation} yields
\begin{align}
  (\M{A} + \M{B}s + \M{C}s^2) \left[ \begin{array}{c} \V{L}^\C{I} \\ \V{t}_\C{B}^\C{A} \\ 1 \end{array} \right] &= \V{0} \notag\\
  (\M{A} + \M{B}s + \M{C}s^2) \V{x} &= \V{0},
  \label{eq:qep}
\end{align}
which stands for the Quadratic Eigenvalue Problem (QEP). QEP is a known problem in Linear Algebra, appeared in connection to dynamical analysis of mechanical systems in \cite{Tisseur-SIAM2001}. There are effective solvers for generalized eigenvalue problems which can be used for solving QEP \cite{Bai-Book2000}. QEP was for the first time exploited in Computer Vision in \cite{Fitzgibbon-CVPR2001} for simultaneous estimation of fundamental or homography matrix and lens distortion. Since then, many problems in computer vision have yielded a solution via QEP, as is the case here. 

The three design matrices $\M{A}$, $\M{B}$, $\M{C}$ must be square matrices and $\M{A}$ a regular matrix.  This yields the minimal number of points to solve \Eq{eq:qep} be $4$. One can feed all the three equations and more than a minimal number of four points and solve the overconstrained system by
$$
    (\M{A}^\top \M{A} + \M{A}^\top \M{B}s + \M{A}^\top \M{C}s^2) \V{x} = \V{0}.
$$ 
Solving such an overconstrained QEP as a Least Squares solution may be biased, as shown in \cite{Steele-ECCV2006}. The method of \cite{Boutry-SIAM2005} proposes to solve rectangular instead of square QEP to reach higher robustness. As for many other problems we do see an improvement as will be demonstrated on the synthetic data in \Sec{sec:synexp}.

\subsubsection{Solver Conditioning}
\label{sec:leverarm_constraint}
\begin{figure}[t]
	\begin{center}
		\begin{tabular}{c@{\hspace{30mm}}c}
			\scriptsize
			\psfrag{C1}[bl][bl][1][-20]{$\V{C}_2$} 
			\psfrag{C2}[Br][Br][1][-20]{$\V{C}_1$}
			\psfrag{I}{$\C{I}$}  
			\psfrag{L}[r][r]{$\V{L}^\C{I}$}  
			\psfrag{g}[r][r]{$\V{g}$}  
			\psfrag{rho}{$\rho$}
			\psfrag{S}{$\V{S}$}
			\includegraphics[width=0.3\linewidth]{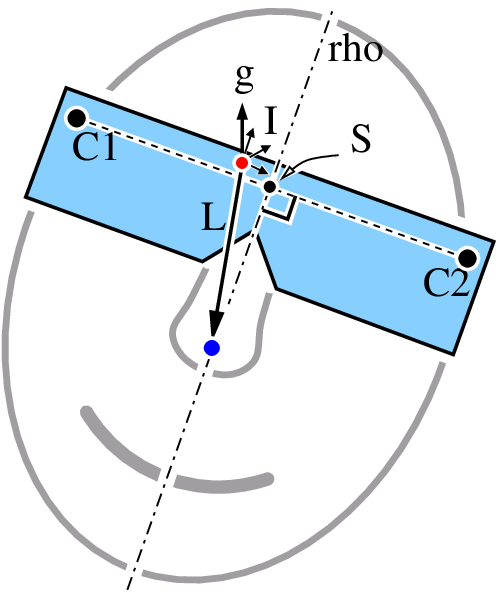} &
			\scriptsize
			\psfrag{C1}[bl][bl][1][-20]{$\V{C}_2$} 
			\psfrag{C2}[Br][Br][1][-20]{$\V{C}_1$}
			\psfrag{L}[r][r]{$\V{L}^\C{S}$}  
			\psfrag{g}[r][r]{$\V{g}$}  
			\psfrag{rho}{$\rho$}
			\psfrag{S}{$\C{S}$}
			\includegraphics[width=0.3\linewidth]{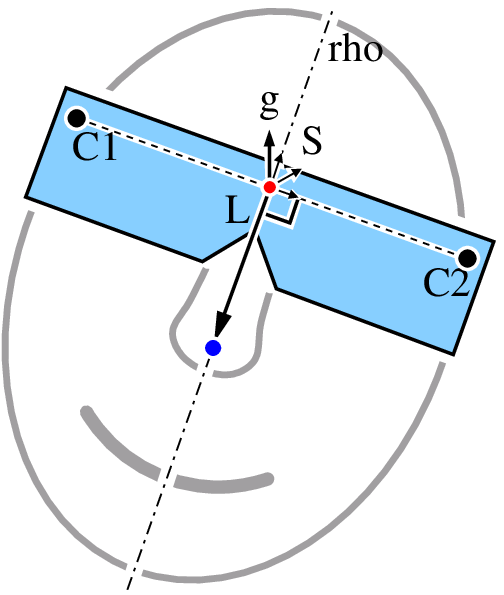} \\
			(a) & (b)\\[-4ex]
		\end{tabular}
	\end{center}
	\caption{In order to well constrain the problem, a 3\MM{d} point on another glasses wearer which we track must be a point on the plane of symmetry $\rho$. Poses between camera $\V{C}_1$, camera $\V{C}_2$, \imu{} $\C{I}$, and mid point $\V{S}$ are all known. Body origin from (a) can be transformed into the mid point $\V{S}$ as shown in (b).}
	\label{fig:head}
\end{figure}

The system in \Eq{eq:qep} is ill posed. The lever arm vector $\V{L}^\C{I}$ and the relative translation $\V{t}_\C{B}^\C{A}$ cannot be uniquely estimated, only up to $1$\MM{dof}. Considering the practical constraints of the targeted use case, this can be remedied. The glasses are in most cases worn such that the middle point between the cameras is also in the center of the head. We can track a point which lies on the face symmetry plane such as a nose or chin, see \Fig{fig:head}(a). This prior knowledge allows to constrain the solver by adding into the matrix $\M{A}$ a linear constraint such that the point $\V{L}^\C{I}$ lies on the plane $\rho$, \ie
$$
[\V{n}^\top d] \left[ \begin{array}{c} \V{L}^\C{I} \\ 1 \end{array}\right] = 0,
$$
with the unit normal $\V{n} = \C{N}(\V{t}_{\C{C}_2}^\C{I}-\V{t}_{\C{C}_1}^\C{I})$ and the distance $d = -\V{n}^\top\,\V{S}$, where $\V{S} = (\V{t}_{\C{C}_2}^\C{I}+\V{t}_{\C{C}_1}^\C{I})/2$ is the mid point between the two cameras.

Alternatively, one can move the origin of the body into the point $\V{S}$ and rotate the bases such that one basis vector is set to $\V{e}_1 = \V{n}$, see \Fig{fig:head}(b). 

The additional constraint on $\V{L}^\C{I}$ can be alternatively enforced as follows.
\begin{itemize}
	\item[{[\MM{c}\oldstylenums{1}]}] Hard constraint on the point $\V{L}^\C{S}$ to lie on the plane of symmetry. The first coordinate of $\V{L}^\C{S}$ is then known to be $0$. It cancels then a column in the design matrices $\M{A}$, $\M{B}$, $\M{C}$ and reduces the unknown vector $\V{x}$ in \Eq{eq:qep} by one element into the final $5$ element vector.
	\item[{[\MM{c}\oldstylenums{2}]}] Soft Constraint on the point $\V{L}^\C{S}$ to lie on the plane of symmetry. One leaves three coordinates of $\V{L}^\C{S}$ but adds the dot product with $\V{n}$ to be $0$ into the design matrices $\M{D}_i$. This allows the solver to relax the strong requirement of the tracked point to  lie on the symmetry plane.
	\item[{[\MM{c}\oldstylenums{3}]}] Hard constraint on the point $\V{L}^\C{S}$ \wrt its prior value $\V{\bar{L}}^\C{S}$. One simply uses the prior value and removes it from the unknown vector. This captures the situation when one tracks a point on the glasses and we a priori know its distance to the \imu{} due to the known 3\MM{d} model of the glasses.
	\item[{[\MM{c}\oldstylenums{4}]}] Soft constraint on the point $\V{L}^\C{S}$ \wrt its prior value $\V{\bar{L}}^\C{S}$. We add additional three rows for each coordinate of the equation  $\bar{\V{L}}^\C{S} - \V{L}^\C{S} = \V{0}$. 
\end{itemize}


With the first two constraints, \ie enforcing the lever arm point to lie on the plane of symmetry, the minimal number of track points is $3$. With the last two constraints, \ie enforcing the lever arms for a given value, either as soft or hard constraint, the minimal number of track points is $2$.

\subsection{Multiple solutions}
There are $14$ solutions of \Eq{eq:qep} for $s$ and $\V{x}$. However, there are $4$ real solutions and typically $2$ of them in a reasonable range. Reasonable range is considered a range which corresponds to angle $\theta$ around $\V{g}$ of $\langle 1, 359 \rangle \,\mathrm{deg}$. Note that $\theta=0$ is the critical configuration, see \Sec{sec:critical_configuration} on critical configuration.

QEP can be be converted into the polynomial in $s$ which might be solved in analytical closed-form way provided we could reduce it to the quartic (or less) degree polynomial. The polynomial comes from the fact that the existence of a non-trivial solution is conditioned by $\det(\M{A} + \M{B}s + \M{C}s^2) = 0$. There is a polynomial of degree $14$, and of degree $12$ when one coordinate of $\V{L}^\C{I}$ is eliminated, respectively. This can be reduced by devision of $s^2+1$ which  would lead to the polynomial of degree $12$ and $10$ respectively. However, if we could eliminate $\V{t}_\C{A}^\C{B}$ from $\V{x}$ to get length of the unknown vector $3$, then the determinant equation would yield quartic polynomial in $s$ which can be solved analytically. For details, see proof in the next.

\subsubsection{Proof}
Let us split the matrix $\M{A} + \M{B}\,s + \M{C}\,s^2$ in \Eq{eq:qep} as
\begin{equation}
[\M{L} \ |\  \M{T} \  | \ \V{l}] \left[ \begin{array}{c} \V{L}^\C{I} \\ \V{t}_\C{B}^\C{A} \\ 1 \end{array} \right] = \V{0},
\label{eq:matrix_split}
\end{equation}
such that the matrix $\M{L}$ multiplies $\V{L}^\C{I}$, the matrix $\M{T}$ multiplies $\V{t}_\C{B}^\C{A}$, and the vector $\V{l}$ multiplies the homogeneous component.
As explained in \Sec{sec:leverarm_constraint}, one degree of freedom of $\V{L}^\C{I}$ can be fixed and so we can assume that the vector $\V{L}^\C{I}$ has only two elements. The matrix $\M{T}$ for minimal number of three tracked points is of the form
$$
\M{T} = \left[ \begin{array}{c} \M{T}_1 \\ \M{T}_2 \\ \M{T}_3 \end{array} \right]
$$
where
$$
\M{T}_i = (1+s^2) \M{R}_i \M{I}_3,
$$
with $\M{R}_i = [\M{R}_{\C{C}_i}^\C{A} \, {\V{p}_i}]_{_\times}$ for $i$th tracked point and $\M{I}_3$ being the 3$\times$3 identity matrix.
We can eliminate $\M{T}$ by left-multiplying the equation \Eq{eq:matrix_split} by the $9 \times 9$ projection matrix
$$
\M{P} = \M{T}(\M{T}^\top \M{T})^{-1} \M{T}^\top.
$$
It can be shown that $s$ fully cancels out from the matrix $\M{P}$ and \Eq{eq:matrix_split} reduces to
$$
\underbrace{[\M{P} \M{L} \ | \ \M{P} \V{l}]}_{\M{G}} \left[ \begin{array}{c} \V{L}^\C{I} \\ 1 \end{array} \right] = \V{0}.
$$
Each consecutive triplet of rows in $\M{G}$ is linearly dependent. If we take only one row from each triplet, $\M{G}$ becomes a $3 \times 3$ matrix. To guarantee a non-trivial solution, it must hold $\det(\M{G}) = 0$. Determinant yields a $6$th degree polynomial in $s$. The polynomial should be divisible by $(1+s^2)$ by definition, see \Eq{eq:basic_equation} which would reduce to a polynomial of degree $4$. Therefore, the minimal solution reduces to analytical solution of a quartic polynomial with exactly $4$ solutions. $\blacksquare$

\subsection{Critical configuration}
\label{sec:critical_configuration}
Note that 
\begin{eqnarray}
\lim_{\theta\rightarrow 0} s = \lim_{\theta\rightarrow 0} \frac{1}{\tan(\frac{\theta}{2})} = \infty.
\end{eqnarray}
This means that $\M{R}_\C{B}^\C{A}$ cannot be estimated if the local coordinate systems have the two axes other than the gravity axis aligned. However, this cannot happen in practice due to noise presence, while it would require that two persons orient themselves at VIO start in exactly the same way. As a remedy, we can freely rotate one of the local CSs around $\V{g}$ twice and run the minimal solver three times to detect and exclude the critical configuration. Twice should be the results consistent.

\subsection{Bi-directional constraint}
\label{sec:bidir}
So far, we considered that one glasses wearer is tracked by another wearer. This way is solvable, as presented above, however, the noise analysis has shown high sensitivity to the noise on image detections and tracks. We can add constraint from the both sides, while still keeping the unknown relative pose in one direction. Following the derivations for the direction when $\C{A}$ tracks $\C{B}$, yet, we consider that $\C{B}$ tracks $\C{A}$. Then, assuming that $\V{K}^\C{A}$ is the lever arm of the person $\C{A}$,
\begin{equation}
   \V{K}^\C{A} = \M{R}_\C{I}^\C{A} \V{K}^\C{I} + \V{t}_\C{I}^\C{A}.
\end{equation}
Expressing $\V{K}^\C{A}$ in $\C{B}$ reads as
$$
\V{K}^\C{B} = \M{R}_\C{A}^\C{B} \V{K}^\C{A} + \V{t}_\C{A}^\C{B} = {\M{R}_\C{B}^\C{A}}^\top \V{K}^\C{A} - {\M{R}_\C{B}^\C{A}}^\top \V{t}_\C{B}^\C{A}.
$$
The point $\V{K}^\C{B}$ projects into camera $\C{A}$ such that
\begin{equation}
  \lambda \hat{\V{p}} = \M{R}_\C{B}^\C{C} \V{K}^\C{B} + \V{t}_\C{B}^\C{C}
  \label{eq:full_projection_opposite}
\end{equation}
Putting all together, the point $\V{K}^\C{I}$ is transformed into camera $\C{B}$ as
\begin{equation}
  \hat{\V{p}}^\C{C} = \M{R}_\C{B}^\C{C} \left( \rul{{\M{R}_\C{B}^\C{A}}^\top} \left( \M{R}_\C{I}^\C{A} \rul{\V{K}_{\ }^\C{I}} + \V{t}_\C{I}^\C{A} \right) - \rul{{\M{R}_\C{B}^\C{A}}^\top} \rul{\V{t}_\C{B}^\C{A}} \right) + \V{t}_\C{B}^\C{C}.
  \label{eq:projected_rayB}
\end{equation}
Note the subtle difference to \Eq{eq:projected_ray}, as the relative rotation $\M{R}_\C{B}^\C{A}$ appears now in transpose. From \Eq{eq:rot_parametrization}, the transposed relative rotation can be expressed as
\begin{align}
{\M{R}_\C{B}^\C{A}}^\top &= \M{R}_\C{B}^\C{A}(-s)\notag\\
                         &=\frac{1}{1+s^2} \left( 2(\V{g}\V{g}^\top - s [\V{g}]_{_\times}) + (s^2 - 1) \M{I}_3 \right).
\label{eq:rot_parametrization_transposed}
\end{align}
Then, substituting \Eq{eq:rot_parametrization_transposed} into \Eq{eq:projected_rayB} yields
\begin{align}
(1+s^2) \M{R}_\C{C}^\C{B} \, \hat{\V{p}}^\C{C} = &  & \big((2 \V{g}\V{g}^\top - \M{I}) \M{R}_\C{I}^\C{A} \, & \V{K}^\C{I} & %
                                    - (2 \V{g}\V{g}^\top - \M{I}) & \V{t}_\C{B}^\C{A} &%
                                    + (2 \V{g}\V{g}^\top - \M{I}) \, \V{t}_\C{I}^\C{A} + \M{R}_\C{C}^\C{B} \V{t}_\C{B}^\C{C}  &\big) \hspace*{2.5mm} + \notag\\
                                 & & + \big(  -2\, [\V{g}]_{_\times} \M{R}_\C{I}^\C{A} \, & \V{K}^\C{I} &%
                                        + 2\, [\V{g}]_{_\times} & \V{t}_\C{B}^\C{A} & %
				    - 2\, [\V{g}]_{_\times}  \V{t}_\C{I}^\C{A} & \big)\,s + \notag\\
                                 & & + \big( \M{R}_\C{I}^\C{A} \, & \V{K}^\C{I} & %
                                        - & \V{t}_\C{B}^\C{A} &%
                                        + \V{t}_\C{I}^\C{A} + \M{R}_\C{C}^\C{B} \, \V{t}_\C{B}^\C{C} & \big)\, s^{2},%
   \label{eq:basic_equation_opposite}
\end{align}
which can be directly re-written into a matrix form. The vectors $(1+s^2) \M{R}_\C{C}^\C{B} \hat{\V{p}}^\C{C}$ and $\lambda \V{p}$ should be the same, up to the detection and calibration errors. Here, $\V{p}$ is constructed from the back projection of $\V{u}^\C{B}$ which is the observation of $\V{K}^\C{I}$ in camera $\C{B}$, \ie $\V{p} = \C{N}(\M{K}_\C{B}^{-1} \V{u}^\C{B})$. The image point $\V{u}^\C{B}$ is the tracked point on the face or glasses of user $\C{A}$. We eliminate their scales and enforce their directions to align, so
\begin{align}
	\lambda \V{p} \times \hat{\V{p}}^\C{C}  &= 0 \notag\\
\lambda \, \M{R}_\C{C}^\C{B} \V{p} \times (1+s^2) \M{R}_\C{C}^\C{B} \hat{\V{p}}^\C{C}  &= 0 \notag\\
	[\M{R}_\C{C}^\C{B} \, \V{p}]_{_\times}\, (1+s^2) \M{R}_\C{C}^\C{B} \hat{\V{p}}^\C{C} & = 0
	\label{eq:cost_opposite}
\end{align}
This gives us two linearly independent equations, hence two constraints. \Eq{eq:cost_opposite} and \Eq{eq:basic_equation_opposite} yields
\begin{align}
  (\M{A}' + \M{B}'s + \M{C}'s^2) \left[ \begin{array}{c} \V{K}^\C{I} \\ \V{t}_\C{B}^\C{A} \\ 1 \end{array} \right] &= \V{0}, 
  \label{eq:qep_opposite}
\end{align}
which again stands for the Quadratic Eigenvalue Problem, sharing the same $s$ and translation $\V{t}_\C{B}^\C{A}$ from the first direction in \Eq{eq:qep}. This allows to concatenate the matrices $\M{A}$, $\M{B}$, $\M{C}$, and $\M{A}'$, $\M{B}'$, $\M{C}'$ into the matrices $\M{D}_0$, $\M{D}_1$, $\M{D}_2$ and to arrive to a solver such that
\begin{equation}
   (\M{D}_0 + \M{D}_1 s + \M{D}_2 s^2) \left[ \begin{array}{c} \V{L}^\C{I} \\ \V{K}^\C{I} \\ \V{t}_\C{B}^\C{A} \\ 1 \end{array} \right] = \V{0}.
   \label{eq:finalQEP}
\end{equation}
To solve the problem, with the symmetry constraint of \Sec{sec:leverarm_constraint}, having lever arms $\V{K}^\C{I}$, $\V{L}^\C{I}$ with $2$\MM{dof}, the relative translation $\V{t}_\C{B}^\C{A}$ with $3$\MM{dof},, the rotation $s$ with $1$\MM{dof}, one needs $7$ equations. Each correspondence adds $2$ linearly independent constraints, so we need minimally $4$ tracked points where at least one point from each direction.

Adding bi-directional constraints has a high impact on robustness of the solver against noise on the detected and tracked points. The bi-directional constraint allows to handle typical noise level and thus turns the solver practical as will be shown in \Sec{sec:experiments}.

There are again multiple solutions of \Eq{eq:finalQEP}. By removing the imaginary ones, one typically ends with maximum of $4$ solutions. To pick the right solution, we evaluate all of them on the $5$th point and choose the one which yields the smallest error in terms of \Eq{eq:finalQEP}. 

\section{Refinement}
\label{sec:refinement}
The closed-form minimal solver builds on the assumption that one possesses perfect VIO ego-poses. This is a reasonable assumption for the minimal solver in order to get an initial estimate. In the next we propose a refinement method which relaxes this assumption and can slightly change the camera poses. The refinement is cast as a non-linear optimization on the image re-projection error as estimating marginal probabilities per unknown node in a Factor Graph, which approximates the MAP estimator. 

The refinement can be formulated in the standard Bundle Adjustment formulation with, \eg, the popular Levenberg-Marquardt optimizer as a non-linear Least Squares ML estimator~\cite{Hartley-Book2004}. Typical non-linear optimization packages like Ceres~\cite{Agarwal-ceres}, g2o~\cite{g2o}, GTSAM~\cite{Dellaert-FTR2017} use sparse linear algebra towards maximum efficiency, possibly parallelizing on multiple CPU cores. However, wearable smart glasses have limited computational CPU power which is shared with other processes like 6\MM{d} ego-pose tracker, rendering, gesture recognition \etc.  Moreover, the glasses have limited battery energy at disposal. Taking these into account, we propose to deviate from the common solution of the Bundle Adjustment and to follow a different route.

We are interested in the refinement stage which would rather use weaker computing units, but many of them, by couple of magnitudes more than CPU cores. Examples of such computational architectures available on mobile devices are  GPU, FPGA, DSP. We formulate the refinement in distributed manner such that the sensed information is shared only between two computational nodes and spread over in a propagation way. We deviate from the standard centralized approach where the whole entire underlined computational structure is available at one place. 

In this line of thinking, the belief propagation in a factor graph shares exactly this property. We aim at finding marginal probabilities on the camera states in the factor graph. This by local message passing as the continuous loopy Gaussian Belief Propagation (GBP). GBP strategy has been proposed by \cite{Ranganathan-IJCAI2007} to solve Simultaneous Localization and Mapping (SLAM) as the Smoothing and Mapping (SAM) problem. This strategy has been introduced before the deep learning era and wide availability of SW and HW parallel computing resources. Only recently was the idea lifted again by \cite{Davison-arxiv2019,Ortiz2020-CVPR2020} and argued for its practicality advantages, in the context of the Graph processor architecture. GBP can be parallelized by design and executed on the modern low energy multi-core platforms. It has been shown in \cite{Ortiz2020-CVPR2020} that GBP for a typical SLAM problem can bring 40-fold speed-up in comparison to the classic BA. GBP is much slower than the standard Bundle Adjustment, when considering a few core CPU, however, its dominance in speed comes when employing massive parallel computing architectures. The classic BA and the GBP mostly converge to the same solution, considering the sum of re-projection errors.

We show in next how the ego-motion alignment can be cast into a GBP inference in a factor graph. In particular, we describe in detail how to construct the graph and how to infer the solution with a message passing framework.

\subsection{Factor Graph}
\begin{figure}
	\begin{center}
		\scriptsize
    	\psfrag{A}[r][r]{camera $\C{A}$}
    	\psfrag{B}[r][r]{camera $\C{B}$}
		\psfrag{Pos}[l][l]{\parbox{1.2cm}{pose prior \\factor}}
		\psfrag{Od}[tc][c]{\parbox[t][1cm]{1.2cm}{odometry \\[2ex] \mbox{~~}factor}}
    	\psfrag{Det}[l][l]{\parbox{1.2cm}{detection \\factor}}
		\includegraphics[width=0.6\linewidth]{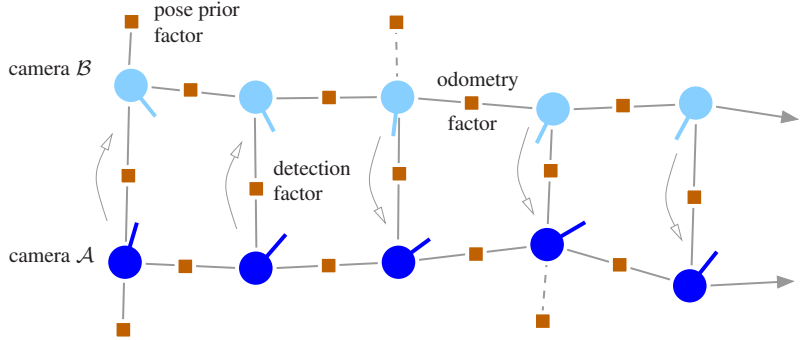}
	\end{center}
	\caption{Factor Graph representing the refinement of the ego-motion alignment. Blue circles represent the two moving cameras, each connected with the odometry factor. When a person $\C{B}$ is seen in camera $\C{A}$, there is an arrow from $\C{A}$ to $\C{B}$ what influences the definition of the factor, but not the message passing. First poses of both cameras are conditioned by the pose prior factors, optionally, any other can be conditioned, depicted by the dashed lines. }
	\label{fig:factorgraph}
\end{figure}

Let us assume that $\C{X} = \{\V{x}_{c} \}, c = 1,2,...,$ is a the set of all the camera poses that we want to refine. The camera poses, and in particular their intra- and inter-camera relations, are constrained by the ego-poses and by the faces or glasses detections, respectively. At the same time, they may be conditioned by prior estimates, \eg estimates from the initial solution. All this can be graphically modeled by an undirected bi-partite graph, a factor graph, with two types of nodes, variable and factor nodes, shown in \Fig{fig:factorgraph}, with the following meaning.
\begin{itemize}
	\item The variables are numerical parameters that are not directly measurable, that is, the parameters we wish to estimate. They represent each camera's poses at times when a face or glasses of the other wearer are detected. The $c$th camera pose stands for $3$-element translation vector and rotation around the gravity vector, $\V{x}_{c} = (x,y,z,\theta)^\top$. 
	\item The factors which join variables represent constraints imposed by measurements from face tracklets and poses from a VIO system. The factors which connect to one variable are measurable priors of the camera poses.
\end{itemize}
	
Such a representation explicitly defines the conditionally independent subsets of variables since each factor node is linked to all the variables nodes on which that factor depends. As a result, the joint distribution over the variables can be easily factorized into a product of functions, one per factor node, that is 
\begin{equation}
p(\C{X}) = \frac{1}{Z}\prod_{f} \psi_f(\V{y}_f),
\label{eq:total_prob}
\end{equation}
where $\psi_f(\V{y}_f)$ is a non-negative potential function defined on subsets of variables, concatenated into a vector $\V{y}_f$, and $Z$ is a normalizing factor that makes $p(\C{X})$ a real density function.

Our goal is the probabilistic inference whereby the parameters are estimated. Such graphs though do not allow an exact inference and approximation schemes must be used. Therefore, we consider a (loopy) belief propagation scheme that pass messages between the nodes in order to compute marginal distributions. In particular, we assume a joint Gaussian distribution and we adopt the GBP scheme for the inference. Note that such an assumption implies i) a Gaussian distribution for any marginal or conditional distribution over subsets of $\C{X}$~\cite{Bishop-MLbook2006} and ii) that finding marginal probabilities (sum-product rule) and the most probable state (max-product rule) is equivalent~\cite{Bickson-PhD2009}.

\subsubsection{Variables}
A variable node represents the camera state, \ie  3\MM{d} pose and 1\MM{d} relative orientation of the VIO local pose to the chosen origin (camera $\C{A}$ in our case). The state vector of camera $c$ is then parameterized by four numbers $\V{x}_c = (x,y,z,\theta)^\top$. Gaussian distribution in the state space can be written as follows
\begin{equation}
p_c(\V{x}_c) = K_c e^{-\frac{1}{2}\big[(\V{x}_c - \bm{\mu}_c)^\top \M{\Lambda}_c (\V{x}_c - \bm{\mu}_c)\big]}
\end{equation}
with mean $\bm{\mu}_c$ and precision, inverse covariance, matrix $\M{\Lambda}_c$. An alternative information form reads
\begin{equation}
p_c(\V{x}_c) = K_c' e ^{\big[-\frac{1}{2} \V{x}_c^\top \M{\Lambda}_c \V{x}_c + \bm{\eta}_c^\top \V{x}_c\big]},
\label{eq:variable_distribution}
\end{equation}
with information vector $\bm{\eta}_c = \M{\Lambda}_c \bm{\mu}_c$. This form has twofold advantage. First, it allows to represent Gaussian distribution with zero information, \ie infinite uncertainty. Second, multiplication of distributions expressed in this form collapses to summing information vectors and precision matrices which is convenient during the inference with the message passing algorithm. 

The goal of the inference is to estimate the marginal probability per variable node which means to iteratively estimate information vectors $\bm{\eta}_c$ and precision matrices $\M{\Lambda}_c$. 

\subsubsection{Factors}
General form of the Gaussian factor is
\begin{equation}
\psi_f(\V{y}_f) = K_f e^{-\frac{1}{2}\big[(\V{h}(\V{y}_f) - \V{z}_f)^\top \M{\Lambda}_f (\V{h}(\V{y}_f) - \V{z}_f)\big]},
\label{eq:factor_orig_distribution}
\end{equation}
where $\V{z}_f$ is the measurement vector, $\M{\Lambda}_f$ is the precision or inverse covariance matrix of that measurement, and $\V{h}(\V{y}_f)$ is a vector valued functional which models the dependence of the involved local states in $\V{y}_f$ to the measurements. The normalization constant $K_f$ does not affect the estimate and can be neglected. In order to plug this factor into the framework along with the variable probability distribution in \Eq{eq:variable_distribution}, one needs to transform the factor from \Eq{eq:factor_orig_distribution} into the Gaussian which is defined on the state vectors $\V{x}_c$
\begin{align}
\psi_f(\V{y}_f) & = K_f e^{-\frac{1}{2}\big[(\V{y}_f - \bm{\mu}_f)^\top \M{\Lambda}'_f (\V{y}_f - \bm{\mu}_f)\big]}\\
             & = K_f' e^{\big[-\frac{1}{2} \V{y}_f^\top \M{\Lambda}'_f \V{y}_f + \bm{\eta}_f^\top \V{y}_f\big]}.
\end{align}
Note that size of $\V{y}_f$ varies depending on the type of the factor. In case of the pairwise factor, $\V{y}_f$ is composed of two state vectors of the two linked variable nodes, \ie $\V{y}_f = [\V{x}_{f_1}^\top \,\,\V{x}_{f_2}^\top]^\top$. The functional $\V{h}(\V{y}_f)$ can be linearized at $\V{\bar{y}}_f$ as 
\begin{equation}
\V{h}(\V{y}_f) \approx \V{h}(\V{\bar{y}}_f) + \M{J}_f(\V{y}_f - \V{\bar{y}}_f),
\label{eq:linearization}
\end{equation}
with $\frac{\partial \V{h}}{\partial \V{y}_f}|_{\V{y}_f=\V{\bar{y}}_f}$
which yields the following parameters of the desired approximated Gaussian
\begin{align}
\bm{\eta}_f & = \M{J}_f^\top \M{\Lambda}_f \Big(\M{J}_f \V{\bar{y}}_f + \V{z}_f - \V{h}(\V{\bar{y}}_f)\Big)\\
\M{\Lambda}_f' & = \M{J}_f^\top \M{\Lambda}_f \M{J}_f.
\end{align}
Full derivation can be found in \cite{Davison-arxiv2019}.

The goal during the message passing inference is to keep updating the information vectors $\bm{\eta}_f$ and precision matrices $\M{\Lambda}'_f$, and send them to the adjacent variable nodes. The re-linearization in \Eq{eq:linearization} may look as additional computational burden, but can be done less frequently than each iteration.

For the investigated ego-motion alignment problem, we have three types of factors
\begin{enumerate}
	\item {\bf Pose-prior factor}. This factor anchors a camera to its prior pose with provided confidence. We set this prior for the first pose of camera $\C{A}$ very strongly and much loosely for camera $\C{B}$ as only this camera is affected by the estimated initial relative pose. The functional $\V{h}(.)$ from \Eq{eq:factor_orig_distribution} reads as
	\begin{equation}
	  \V{h}(\V{y}_f) = \V{h}(\V{x}_f) = \left[ \V{x}_f\right].
	  \label{eq:h_poseprior}
	\end{equation}
	Note that $\V{y}_f$ contains just one variable, hence collapses to $\V{x}_f$. The measurement vector $\V{z}(.)$ comes from VIO and the initial estimate for the relative transformation. For camera $\C{B}$ it is set as
	\begin{equation}
	  \V{z}_f(\V{x}_c) = %
	  \left[ \begin{array}{c} \V{t}_\C{B}^\C{A} + \M{R}_\C{B}^\C{A}\, \V{t}_\C{I}^\C{B} \\ %
	  \theta(\M{R}_\C{B}^\C{A})
	   \end{array}\right],
	\end{equation}
	where $\theta(\M{R}_\C{B}^\C{A})$ is the pan angle read out from the estimated initial rotation matrix $\M{R}_\C{B}^\C{A}$ of the closed-form solver in \Eq{eq:finalQEP}. Recall that the matrix is constructed from the $s$ parameter in \Eq{eq:finalQEP} which can be converted to an angle. The confidence of the pose prior is controlled by the precision matrix and set to
	\begin{equation}
	\M{\Lambda}_f = \left[ \begin{array}{cc} \frac{1}{\sigma_\V{t}^2} \M{I}_3 & \\%
	& \frac{1}{\sigma_\theta^2}\end{array}\right],
	\label{eq:precision_matrix}
	\end{equation}
	with $\M{I}_3$ being the $3\times3$ identity matrix, and $\sigma_\V{t}$ and $\sigma_\theta$ the standard deviations on translation and angular part, respectively. These are set according to our confidence on the provided initial estimation of the relative pose $\M{R}_\C{B}^\C{A}$ and $\V{t}_\C{B}^\C{A}$. The functional $\V{h}(.)$ in \Eq{eq:h_poseprior} is linear in state variables, hence $\M{J}_f = \M{I}_4$.
	
	\item {\bf Intra-camera factor}. This factor links the same camera at different times and is called the odometry factor. For instance, for camera $\C{B}$ it reads as
	\begin{equation}
	\V{h}(\V{y}_f) = \left[\begin{array}{c}%
						 \M{R}_\C{A}^\C{B}(\V{y}_{f,\{4\}}) (\V{y}_{f,\{5:7\}} - \V{y}_{f,\{1:3\}}) \\
	                     \V{y}_{f,\{8\}} - \V{y}_{f,\{4\}} %
	                  \end{array} \right] = %
 					  \left[\begin{array}{c}%
					  \M{R}_\C{A}^\C{B}(\V{x}_{f_1,\{4\}}) (\V{x}_{f_{2},\{1:3\}} - \V{x}_{f_{1},\{1:3\}}) \\
					  \V{x}_{f_{2},\{4\}} - \V{x}_{f_{1},\{4\}} %
					  \end{array} \right] 
	\end{equation}
    with 
	\begin{equation}
	  \newcommand{\myang}{\theta}
	  \M{R}_\C{A}^\C{B}(\myang) = %
	  \left[\begin{array}{ccc}%
	  \cos(\myang) & -\sin(\myang) & 0 \\ \sin(\myang) & \cos(\myang) & 0 \\ 0 & 0 & 1
	  \end{array}\right],
    \end{equation}
    where $\V{y}_{f,\{5:7\}}$ stands for a 3-vector composed of 5, 6, 7th element from the entire 8-element vector $\V{y}_f$.
    For the odometry, we employ VIO ego-poses of the camera at two neighboring poses with indices $f_1$ and $f_2$. The measurement vector $\V{z}_f$ is defined as  
	\begin{equation}
		\V{z}_f = \left[\begin{array}{c}%
		\V{t}_\C{I}^\C{B}(f_2) - \V{t}_\C{I}^\C{B}(f_1) \\
		0
		\end{array} \right], 
	\end{equation}
	where, to recall, $\V{t}_\C{I}^\C{B}(f_2)$ is the translation of camera $\C{B}$ at $f_2$ in the local coordinate system with origin $\C{O}_\C{B}$, see \Fig{fig:motivation}. The measurement precision matrix is set as in \Eq{eq:precision_matrix} %
    except that $\sigma_\V{t}$ and $\sigma_\theta$ are the standard deviations on translation part of VIO ego-pose (set to $5$mm per $1$m) and on its angular part (set to $0.1$deg per $1$m), respectively. These values represent certainty of a common VIO system, \eg \cite{Mourikis-ICRA2007}. The partial derivatives in the $3\times8$ Jacobian $\M{J}_f$ \wrt the translational part are constant, so only the two columns of the rotational part need linearization. Derivation of the Jacobian is straightforward and omitted.
	
	\item {\bf Inter-camera factor}. This factor links different cameras and is called the image detection factor. The constraints come from face or glasses tracklets expressed via the re-projection error as explained in the following. For camera $\C{B}$ in which person $\C{A}$ is tracked it reads as follows
	
	\begin{equation}
	\V{h}(\V{y}_f) = %
	\left[ \C{H} \bigg( \M{K}_\C{B} \Big( \M{R}_\C{B}^\C{C} \, \M{R}_\C{A}^\C{B}(\V{y}_{f, \{4\}})\,\V{p^\C{A}} + \V{t}_\C{I}^\C{C} \Big) \bigg) \right] ,
	\end{equation}
	where $\C{H}(.)$ returns an inhomogeneous 2 coordinate vector and
	\begin{equation}
	   \V{p^\C{A}} = \V{x}_{f_2,\{1:3\}} - \V{x}_{f_1,\{1:3\}} + \M{R}_\C{I}^\C{A} \, \V{K}^\C{I}.
	\end{equation}
	Camera $\C{B}$ is indexed by $f_1$, while camera $\C{A}$ by $f_2$, and we plot this into the factor graph as the direction from $\C{B}\rightarrow \C{A}$ in \Fig{fig:factorgraph}. 
	The measurement $\V{z}_f(.)$ is the detection of the tracked face or glasses point and is represented as two image coordinates in camera $\C{B}$ as
	\begin{equation}
	   \V{z}_f = \left[ \V{u}^\C{B} \right]
	\end{equation}
	with $\M{\Lambda}_f = \frac{1}{\sigma_\V{u}^2}\M{I}_2$, where $\sigma_\V{u}$ is the standard deviation of the detection in pixels, set to $1$\,pixel in our experiments. It reads analogous for camera $\C{A}$ which sees the person $\C{B}$. The $3 \times 8$ Jacobian $\M{J}_f$ is again constant \wrt the translation part of the state vectors and a non-linear function of the angular part. Derivation of the Jacobian is straightforward and omitted.
	
	
\end{enumerate}

\subsubsection{Gaussian Belief Propagation}

In GBP, messages have the form of the probability distribution which is expressed by information vectors $\bm{\eta}_c$, $\bm{\eta}_f$ and precision matrices $\M{\Lambda}_c$, $\M{\Lambda}'_f$. The informations vectors are iteratively updated and passed to directly connected nodes. This way, GBP  is  able  to  efficiently  determine marginal distributions for every variable in a tree graph with a one  time  forward / backward  sweep  of  message  passing through the graph and reach thus a global optimum. For loopy graphs, like the one for the ego-motion alignment, there is no theoretical guarantee to reach the optimum. In this case the message passing is let to iterate until convergence. It has been widely demonstrated that loopy GBP converges to a meaningful solution in many practical problems. We can confirm this as well for the ego-motion alignment problem.  

Let us emphasize that the refinement has freedom to cope with a typical drift in VIO ego-poses as each state of the variable node in the graph represents absolute camera pose and is directly optimized over. The relative transformation in the refinement stage is implicitly encoded as it is not directly parameterized and tuned. The initial relative transformation only serves to bring user $\C{B}$ into the coordinate system of user $\C{A}$ and this with some pre-set certainty, as mentioned, small for user $\C{A}$ and high for user $\C{B}$. From that moment on, there are only the absolute poses of both users which are being optimized over. 

Treating outliers during GBP can be handled by incorporating a robust, \eg Huber, kernel as shown in \cite{Davison-arxiv2019}. This allows to feed into the system wrong image detections which are result of a failing face detector or image patch tracker. In our experiments, we experimentally verified feasibility of incorporating directly the robust Huber kernel. However, we run the minimal solver in RANSAC loop first in order to get reasonable initial estimation of the relative transformation. This prunes the outliers and are then not passed into the GBP.
\section{Face and Glasses Tracker}
\label{sec:faceGlassesTracker}
In \Sec{sec:leverarm_constraint}, we discussed several constraints that can be optionally enforced. In fact, such constraints depend on the underlying point tracker. When facial landmarks are tracked, $\V{L}^\C{I}$ and $\V{K}^\C{I}$ have to be estimated too. Instead, when glasses points are tracked and the 3\MM{d} model of the glasses with hardware components is also provided, their prior values are available.

\subsection{Face tracking}
Any robust-to-occlusion face tracker that provides facial landmarks can be employed for this task. Even very simple constellation models include facial landmarks on the vertical axis of the face, \eg central nose or lip landmark. One, however, should take into account the landmark tracking inaccuracy because of the non-rigid nature of the face appearance and the occluded eye area from the glasses\footnote{Bulky AR glasses may occlude a large area of the face top.}. Therefore,  constraints [\MM{c}\oldstylenums{1}] or [\MM{c}\oldstylenums{2}] in \Sec{sec:leverarm_constraint} must be adopted in this case. Note that frame-wise tracking is not necessary as one would typically use observations from some key-frames, that is, face detection with landmark localization for specific frames may suffice. The OpenCV face detector based on the Single Shot Detection method~\cite{Liu-ECCV2016} is employed. This algorithm was experimentally validated to be robust enough to outliers such as glasses and sufficiently serves for needs of the ego-motion alignment. 

\subsection{Glasses tracking}
\label{sec:glasses_tracing}
A more accurate tracking that leads to a less sensitive solver can be achieved when the glasses are localized on faces, and their points are tracked. This is due to a series of advantages that arise when glasses points instead of facial landmarks are tracked.
\begin{itemize}
  \item AR glasses are rigid devices and can be tracked as single bodies.
  \item When the 3\MM{d} model of the glasses is available, the position of the glasses surface points (front-side perimeter, skeleton, branches \etc) in the \imu{} frame is a priori known.
  \item $\V{L}^\C{I}$ = $\V{K}^\C{I}$, up to manufacturing error and constraints [\MM{c}\oldstylenums{3}] or [\MM{c}\oldstylenums{4}] from \Sec{sec:leverarm_constraint} are to be enforced. 
  \item The front side can be usually approximated by a single planar or multi-plane segments, which allows for 2\MM{d} modeling. Instead, a 3\MM{d} pose tracker \cite{Crivellaro-ICCV2015,Tekin-CVPR2018} can be trained for more complicated form factors. 
\end{itemize}

As with the Snap Spectacles, the majority of AR glasses tend to resemble the standard shape of eye-glasses, where the front side is approximately planar. This favours the use of a 2\MM{d} tracker. Therefore, and without loss of generality, we here consider a homography tracker, which can be easily extended to a multi planar tracker. 

A model-to-frame approach is adopted, where a \emph{binary} image model of the front side of glasses is frame-wise aligned against glasses instances. Note that the front side of glasses may appear slanted or be partially occluded. In our case, the binary model becomes immediately available from rendering the front side of the Spectacles 3\MM{d} model. When such a model is not available, a similar mask can be either learnt from a few aligned images or get extracted by a single image that shows the front side of the glasses. To solve the alignment problem, our tracker builds on the ECC algorithm~\cite{Evangelidis-PAMI2008}. However, in order to make the tracker more robust and accurate, we track only the perimeter of the front side, that is, ECC is using only the intensities of the points around the model perimeter. The initialization of the tracker can benefit from a face detector, since the search area for the first track is bounded by the face box.

\begin{figure}
	\if\internalversion1
		\begin{tabular}{c@{\hspace{1mm}}c}
		  \includegraphics[width = 0.45\linewidth]{\Figs hermosa_front_side_rendering} &
		  \includegraphics[width = 0.45\linewidth]{\Figs perimeter_mask}\\
		\end{tabular}
		\caption{The binary image model and the area around the perimeter used by the ECC tracker.}
	\else
		\begin{tabular}{c@{\hspace{1mm}}c@{\hspace{1mm}}c}
	  		\includegraphics[width = 0.32\linewidth]{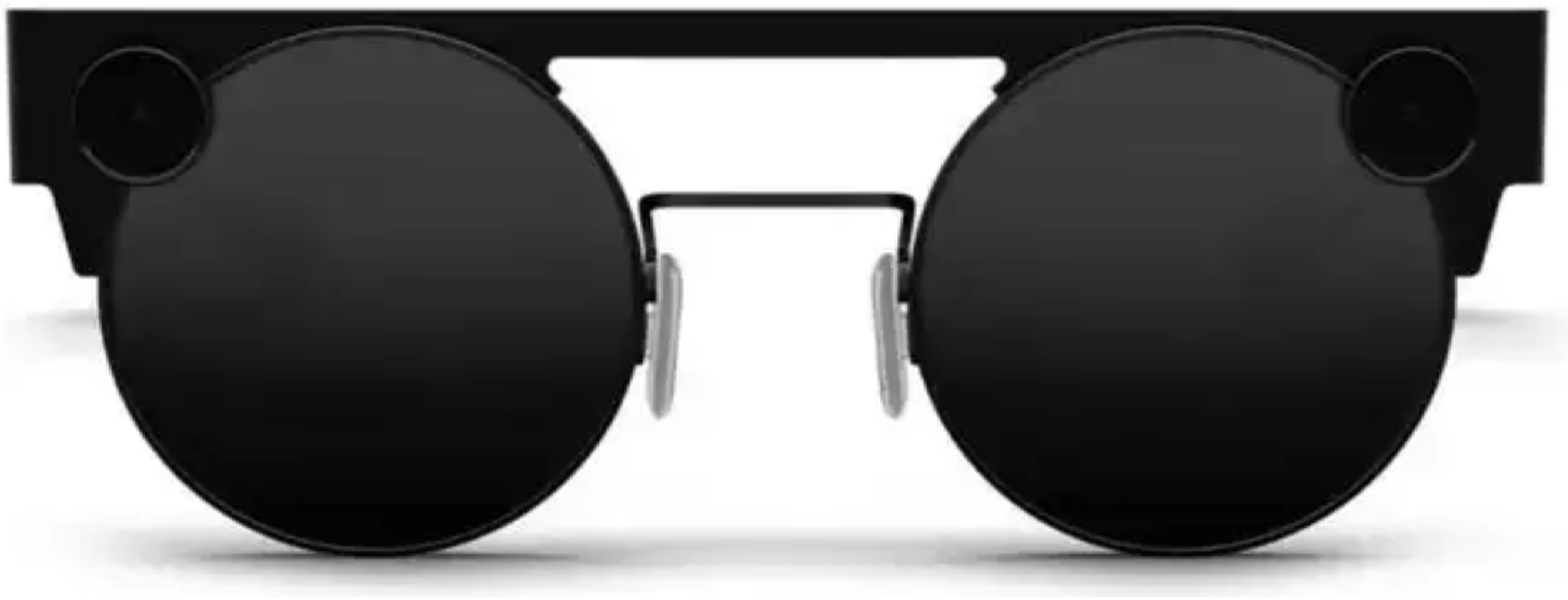} &
		  	\includegraphics[width = 0.32\linewidth]{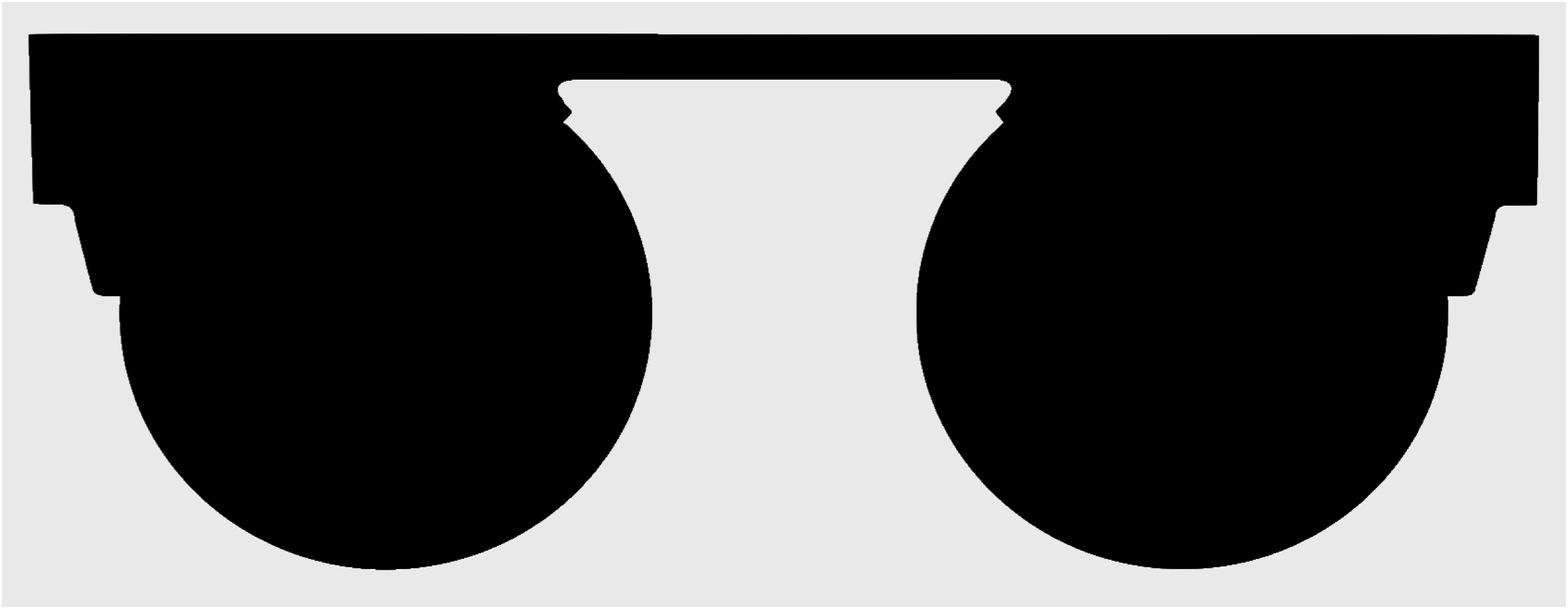} &
		  	\includegraphics[width = 0.32\linewidth]{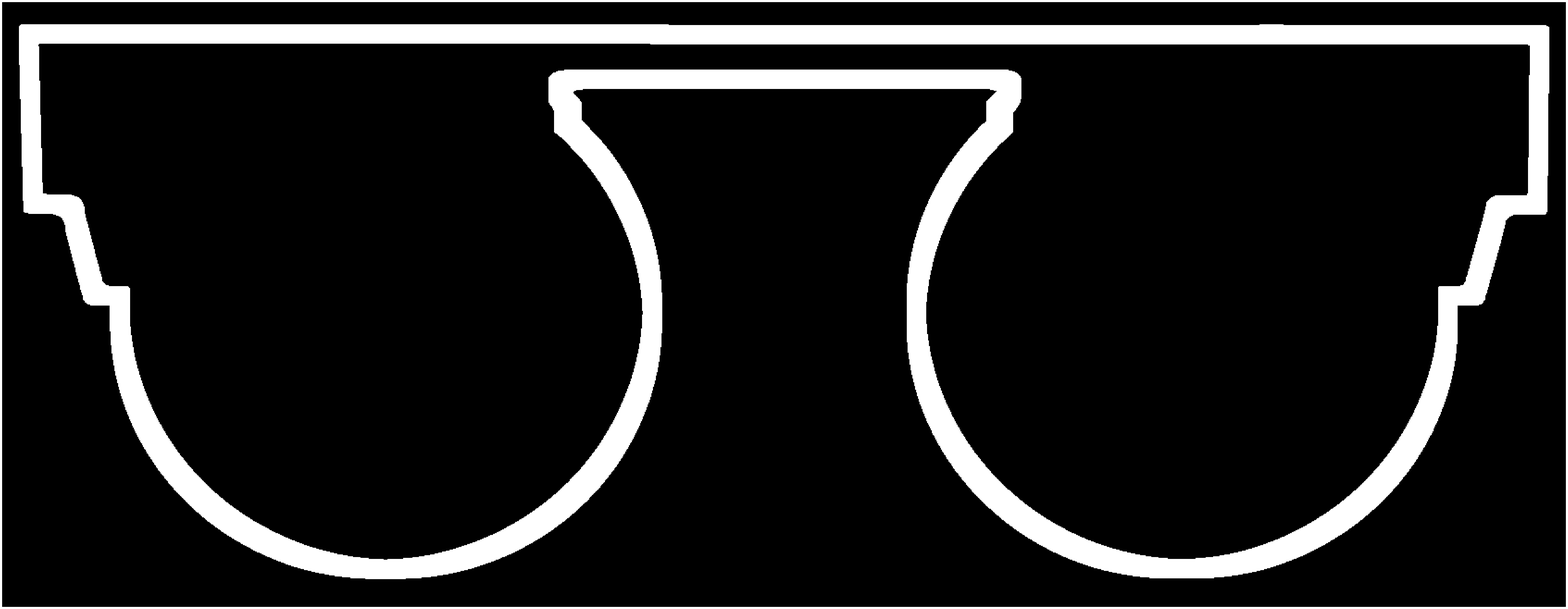}\\
		\end{tabular}
		\caption{The Snap Spectacles 3 glasses, their binary image model and the area around the perimeter used by the ECC tracker.}
	\fi

	\label{fig:glasses_mask_and_perimeter}
\end{figure}

\Fig{fig:glasses_mask_and_perimeter} shows the rendering mask of the front side of the glasses 3\MM{d} model, which plays the role of the image model, and the support area around the perimeter that is used by the ECC algorithm.

\subsubsection{Formulation}
In short, let us assume a set of 2\MM{d} points $\M{X}=[...,\V{x}_j,...]$, $j=1,...,N$ that mark the model perimeter area, where $\V{x}_j = [x_j, y_j, 1]^\top$ denotes a $2$\MM{d} homogeous point. The ECC algorithm aims at maximizing the correlation between the model appearance $J(\M{X})$ and the warped image $I_k(\M{H}_k \, \M{X})$ of the $k$-th frame, whereby the homography $\M{H}_k$ that aligns the images is estimated. Given the next frame, $\M{H}_{k+1}$ is initialized by $\M{H}_{k+1} = \M{G}_{k}^{k+1} \M{H}_k$, where $\M{G}_{k}^{k+1}$ is a translation transformation from $I_k$ to $I_{k+1}$, namely, the displacement of glasses area from frame to frame. This transformation is estimated by a patch matching step whereby the surrounding area of last glasses localization is scanned. Then, $\M{H}_{k+1}$ is refined using a multi-resolution version of the ECC algorithm, and the process is repeated per frame. The initialization step of our tracker uses a face detection result, that is, the face area is localized in the first frame, a low-resolution patch matching scheme initializes the translation part of homography $\M{H}_0$ by localizing the central-bottom part of the glasses, and $\M{H}_0$ is finally refined by ECC. The same steps are followed whenever re-initialization is needed.

\if\internalversion1
Fig.~\ref{fig:glasses_tracking_examples} shows instances of perimeter tracking. The images correspond to $145\times 130$ cropped images from VGA frames.
\begin{figure}
	\begin{tabular}{c@{}c@{}c@{}c@{}c}
		\includegraphics[width = 0.195\linewidth]{\Figs a4} &
		\includegraphics[width = 0.195\linewidth]{\Figs a2} &
		\includegraphics[width = 0.195\linewidth]{\Figs a8} &
		\includegraphics[width = 0.195\linewidth]{\Figs a7} &
		\includegraphics[width = 0.195\linewidth]{\Figs a5} \\

		\includegraphics[width = 0.195\linewidth]{\Figs b14} &
		\includegraphics[width = 0.195\linewidth]{\Figs b6} &
		\includegraphics[width = 0.195\linewidth]{\Figs b8} &
     	\includegraphics[width = 0.195\linewidth]{\Figs b11} &
		\includegraphics[width = 0.195\linewidth]{\Figs b4} \\
	\end{tabular}
	\caption{Glasses tracking instances: the perimeter of the front side is tracked using the ECC algorithm.}
	\label{fig:glasses_tracking_examples}
\end{figure}
\fi
\section{Experiments}
\label{sec:experiments}
First, we provide perturbation noise analysis on synthetic data. Second, we show results from real experiments with Snap Spectacles glasses on standard use cases when users stand or sit in front of each other and collaborate on a virtual object placed in between. 
\subsection{Synthetic data}
\label{sec:synexp}
In this section, we present perturbation analysis to investigate sensitivity of the solvers \wrt the point tracking noise. 

\begin{figure}[t]
  \begin{center}
    \scriptsize
    \psfrag{x [m]}{$x$[m]}
    \psfrag{y [m]}[c][c]{$y$[m]}
    \psfrag{z [m]}[c][c]{$z$[m]}
    \psfrag{1}[b][t]{$1$}
    \psfrag{2}[b][t]{$2$}
    \psfrag{3}[b][t]{$3$}
    \psfrag{4}[b][t]{$4$}
    \psfrag{5}[b][t]{$5$}
    \psfrag{6}[b][t]{$6$}
    \includegraphics[width=0.8\linewidth]{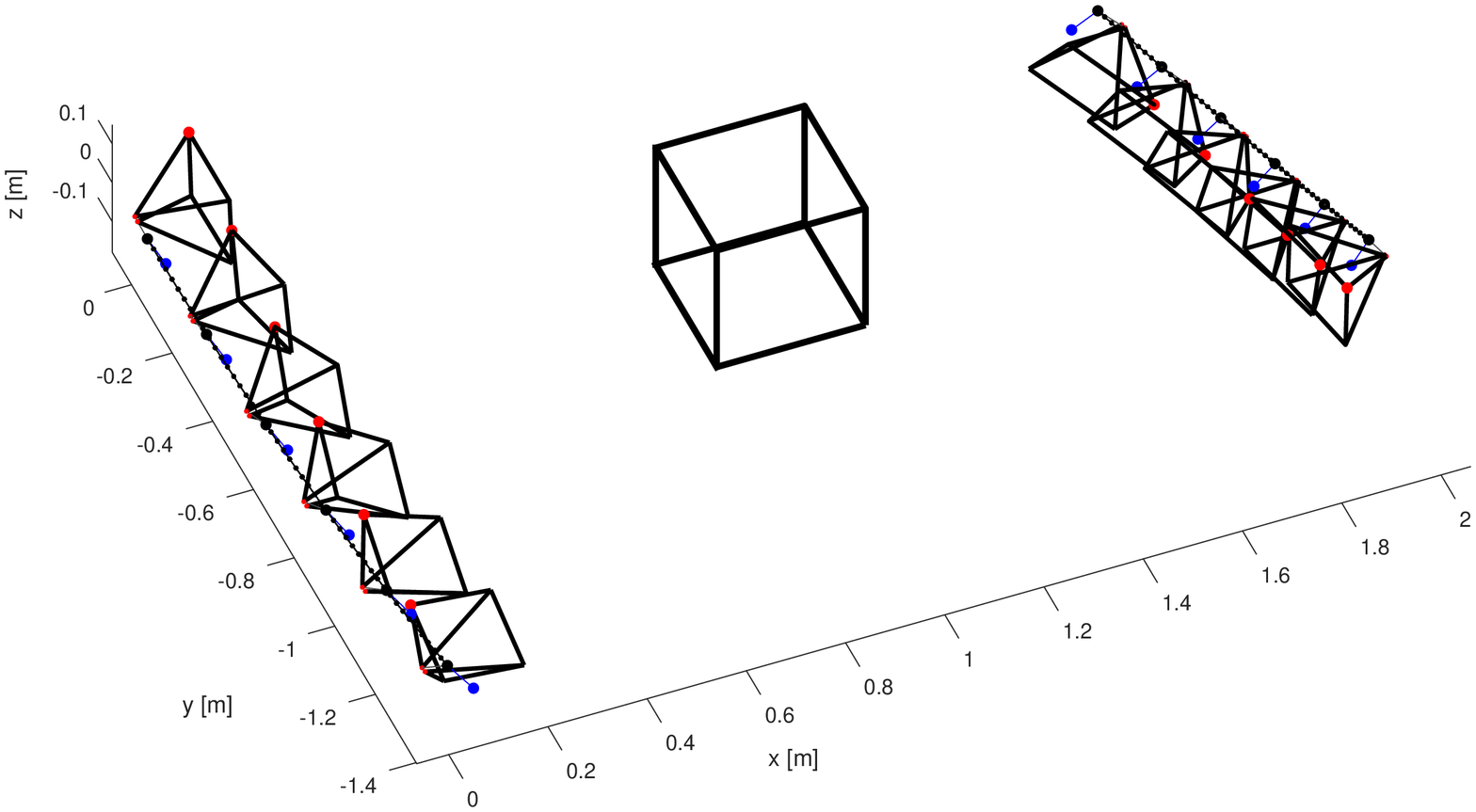} \\[4ex]
    \begin{tabular}{cc}
      \includegraphics[width=0.35\linewidth]{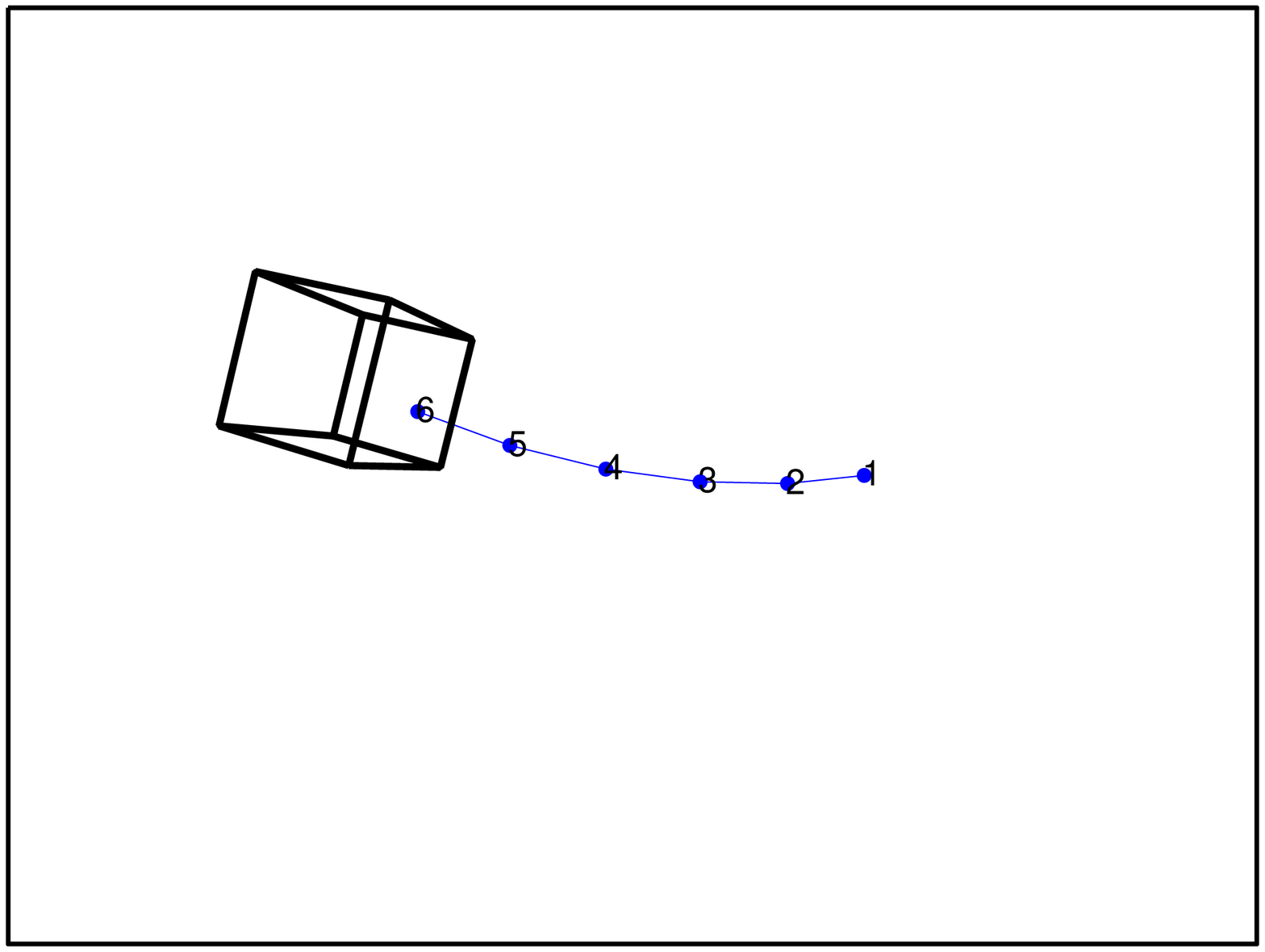} &
      \includegraphics[width=0.35\linewidth, clip=true]{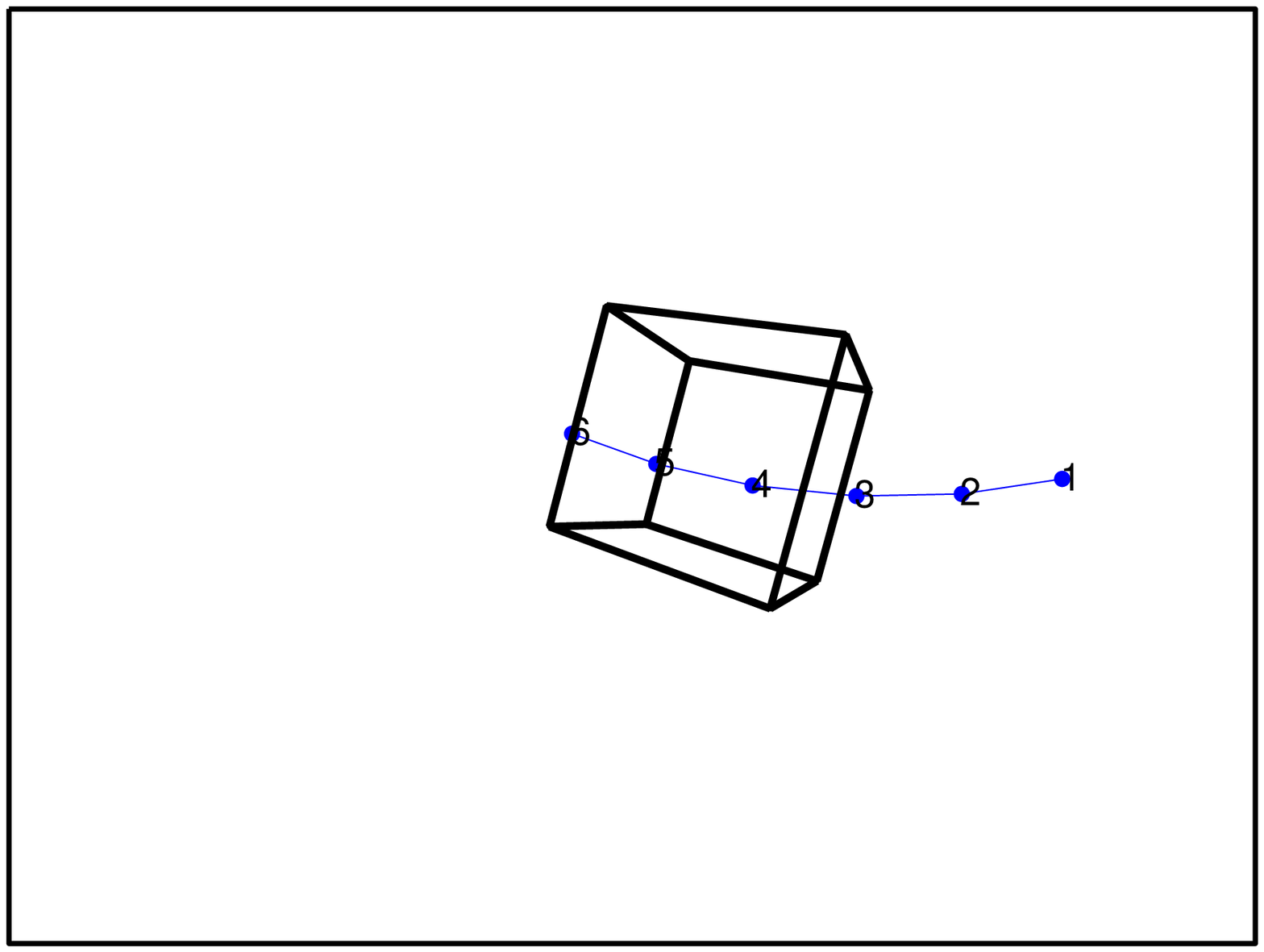} 
    \end{tabular}
  \end{center}
  \caption{Camera setup. Top: Two rolling shutter cameras move and their locations at six time instances are shown. There are 3\MM{d} lever arm points which are tracked in the counter camera. The two lever arms are unknown but rigid to the local CSs of the moving cameras, shown as blue dots. Bottom: Views of the two cameras as they observe the counter lever arm points. A cube projection at the last position of the camera is shown.}
   \label{fig:camera_setup}
\end{figure}

\begin{figure}[t]
    \begin{tabular}{cc}
      \includegraphics[width=0.48\linewidth]{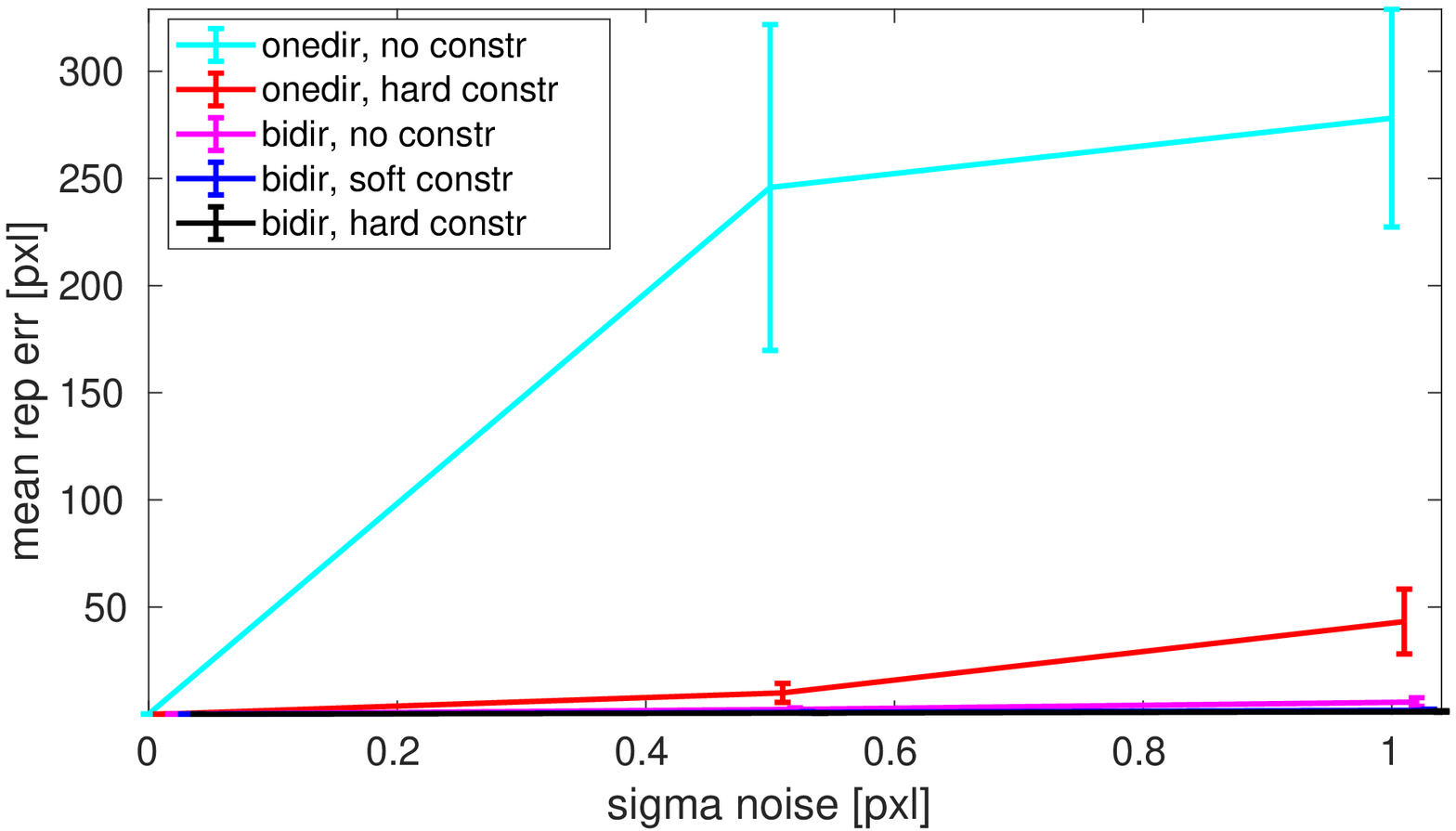} &
      \includegraphics[width=0.48\linewidth, trim = 1 0 0 1, clip=true]{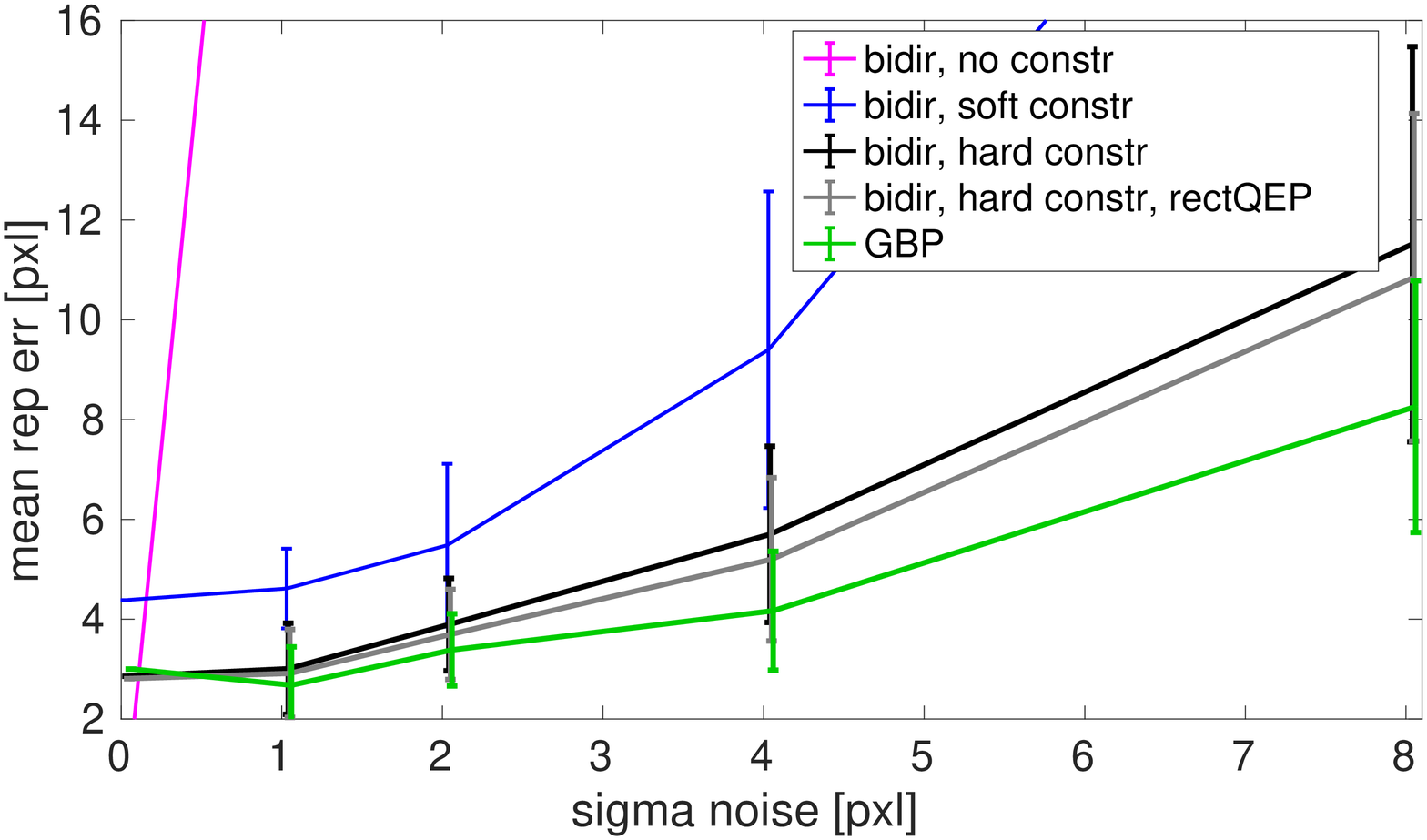} \\
	(a) & (b) \\
    \end{tabular}
  \caption{Mean re-projection error of the projected cube vertices \wrt the noise added on the tracked points. (a) Both one- and bi-directional solvers are shown for noise up to $1$\, pixel sigma. (b) Performance of bidirectional solvers are shown for wider noise range. For four bottom solvers, prior values of the lever arms fed into the solvers were shifted from their ground truth values in all three axes by 10\%. }
  \label{fig:noise_perturbation}
\end{figure}

The setup is a fully synthesized scenario with two moving rolling shutter cameras with lever arms, as shown in \Fig{fig:camera_setup}. The sensitivity of the solvers can be seen in \Fig{fig:noise_perturbation}. As the evaluation metric we compute the mean re-projection error of the eight cube vertices, when projecting them from their $3$\MM{d} locations in the CS of camera $\C{A}$ into the image of camera $\C{B}$ to their ground truth projections by the estimated relative transformation $\M{R}_{\C{B}}^{\C{A}}$, $\V{t}_{\C{B}}^{\C{A}}$. This metric naturally combines inaccuracies in the estimated rotation and translation and gives indication, how much the virtual content would be shifted in the see-through displays of smart glasses from its ideal ``real world'' position. In other words, user $\C{A}$ places virtual scene content, and user $\C{B}$ would see it correctly posed, namely at the correct location and properly oriented, as long as the relative transformation is accurate. 

The perturbation analysis allows us to judge on practicality of the solvers.  \Fig{fig:noise_perturbation}(a) depicts analysis of both one- and bi-directional solvers for small amount of noise, up to $1$ pixel sigma. It can be seen that one directional solvers are too sensitive for any practical use as they deliver satisfactory results only with unrealistically small amount of noise. Bidirectional solvers can handle much more severe noise, as \Fig{fig:noise_perturbation}(b) depicts. Moreover, we deliberately shifted priors on lever arms $\V{K}^\C{I}$, $\V{L}^\C{I}$ from their ground truth values by $10$\%. This captures the real situation such that we track a nose or a point on the worn glasses and our prior on its location to the origin of the \imu{} coordinate system is not perfect. If we do not enforce any prior (labeled as bidir with no constraint method), pixel noise with sigma $0.5$ yields inaccuracy up to $10$ pixels. This is theoretically interesting, but practically acceptable only in case when the tracker is very accurate. However, putting the prior as a soft or a hard constraint keeps the solution in reasonable bounds for much higher noise. The hard constraint is the safest option and the most preferable solution. As a result, only bidirectional solvers can handle typical range of noise on the image points in real applications. As expected, GBP delivers superior solution, which refines on top of the closed form solver with rectangular QEP. For all the solvers, we run them in the overconstrained setup with all $6$ points.

\subsection{Real data}

\begin{figure}
	\begin{tabular}{c@{\hspace{1mm}}c}
		\includegraphics[width = 0.49\linewidth]{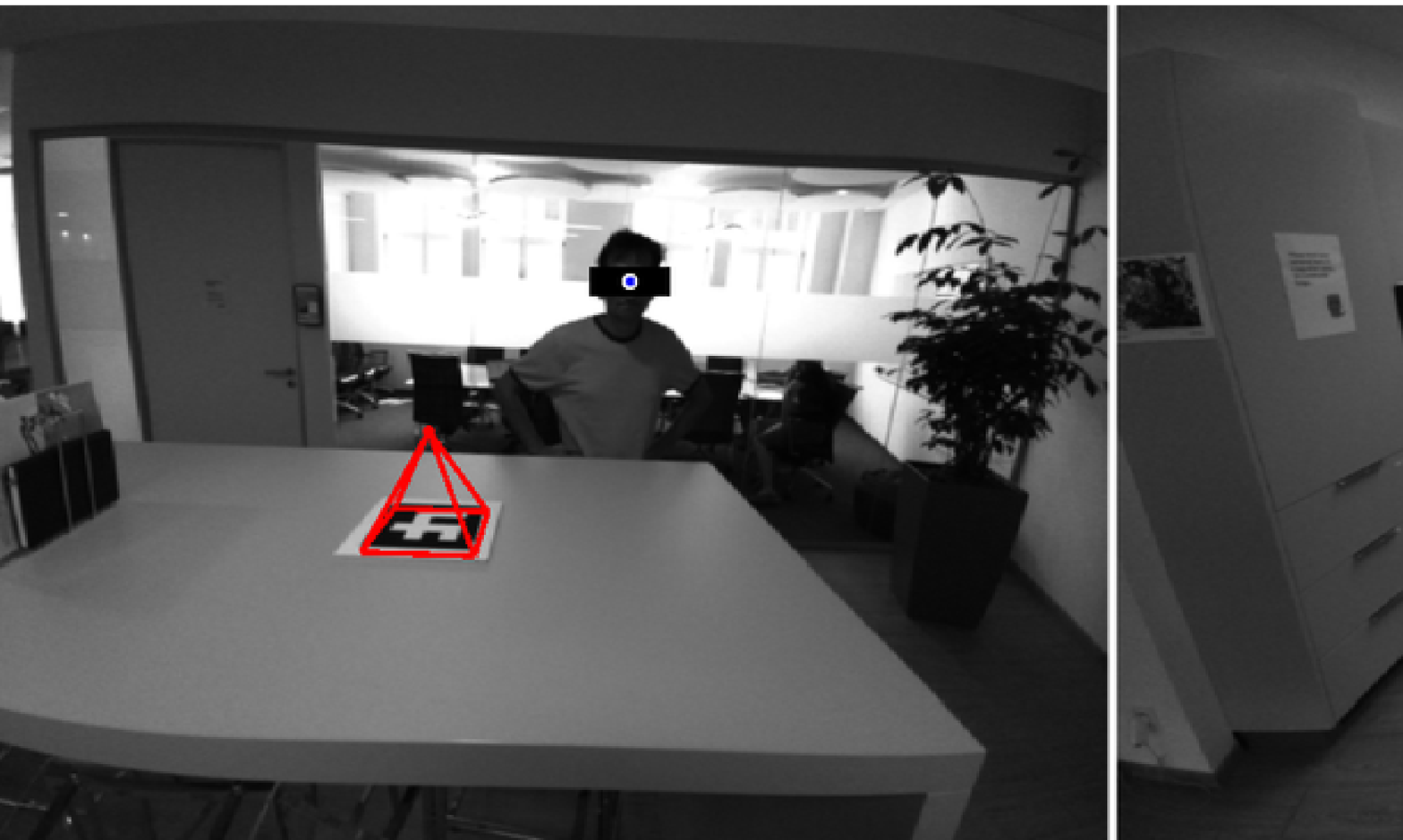} &
		\includegraphics[width = 0.49\linewidth]{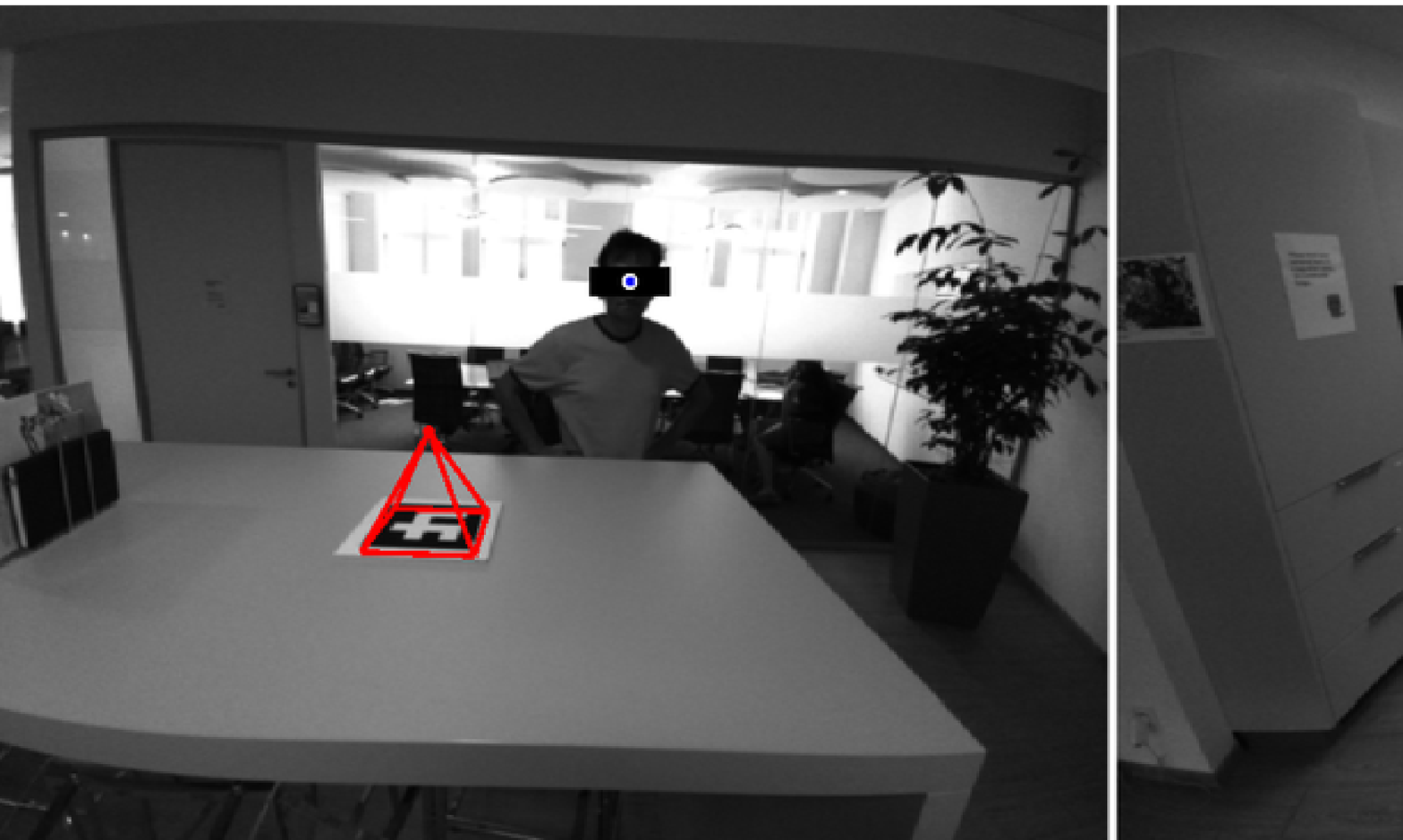} \\
		\includegraphics[width = 0.49\linewidth]{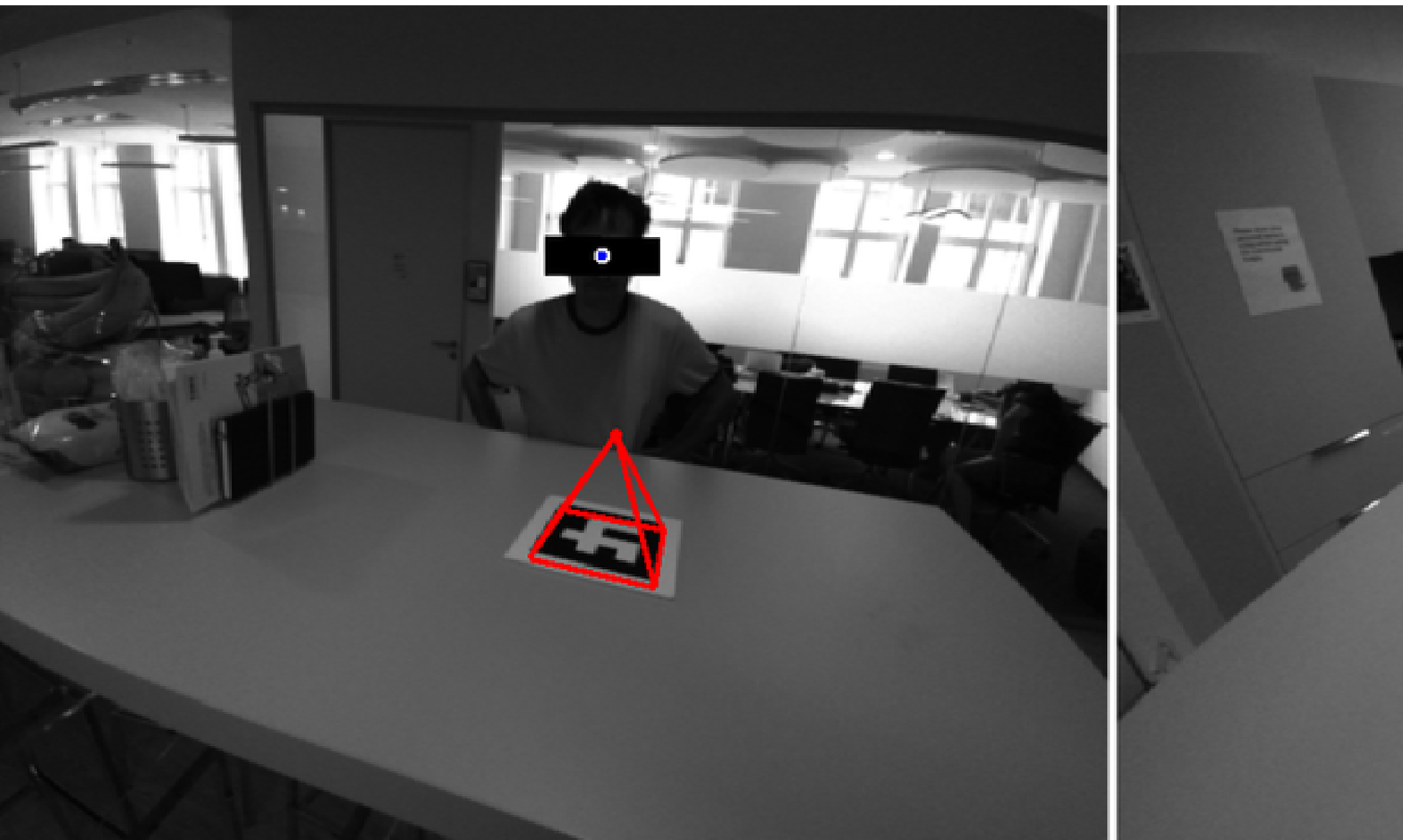} &
		\includegraphics[width = 0.49\linewidth]{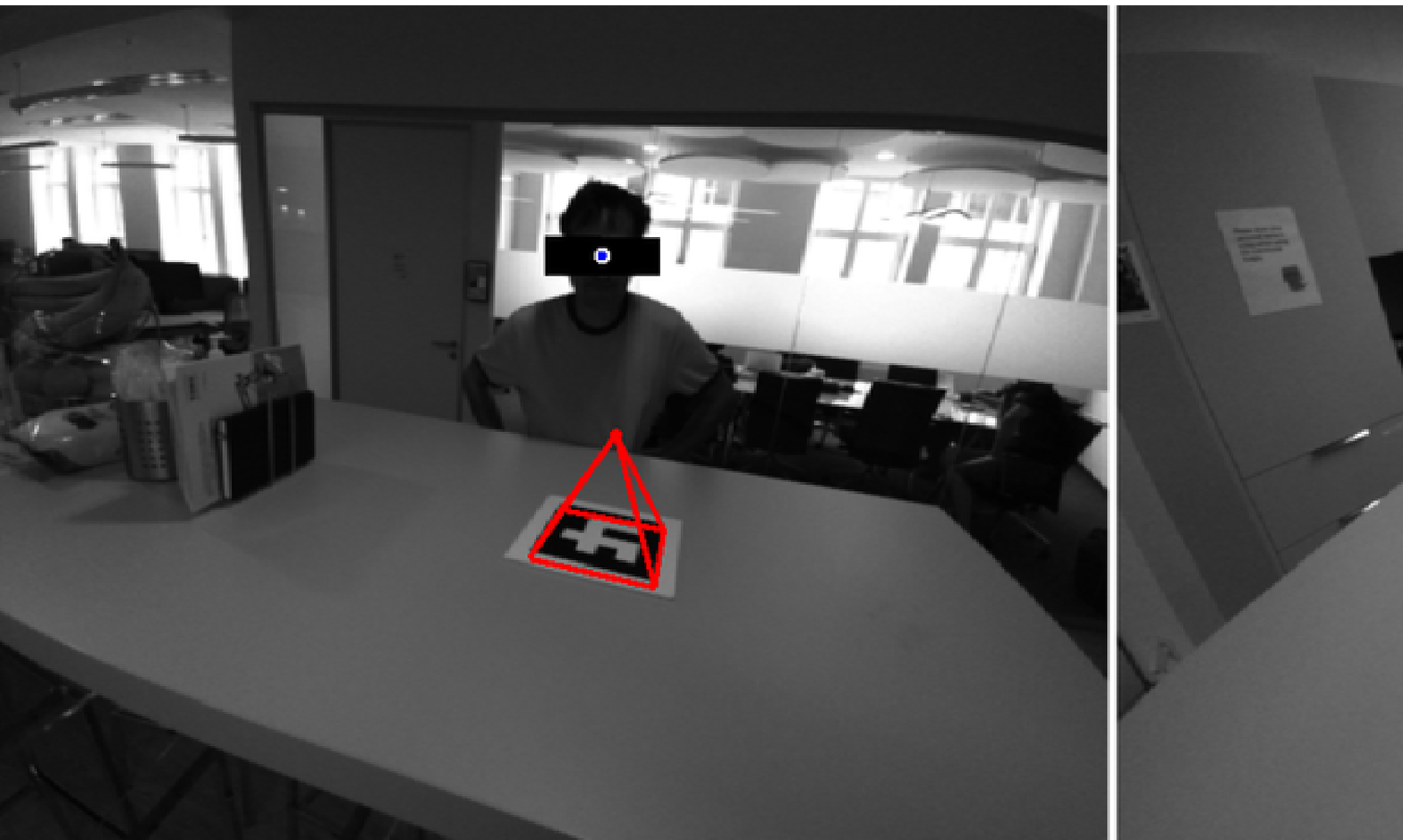} \\[-1ex]
		minimal solver & refinement \\[2ex]
		\includegraphics[width = 0.49\linewidth]{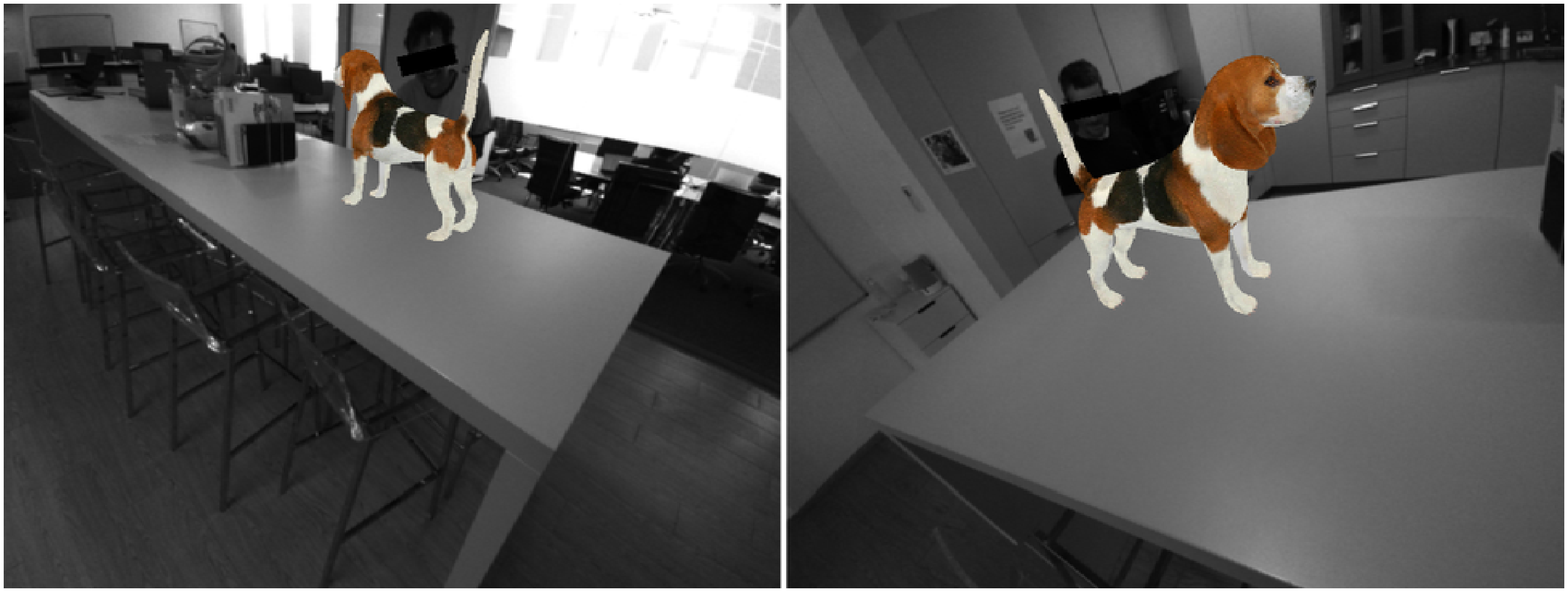} &
		\includegraphics[width = 0.49\linewidth]{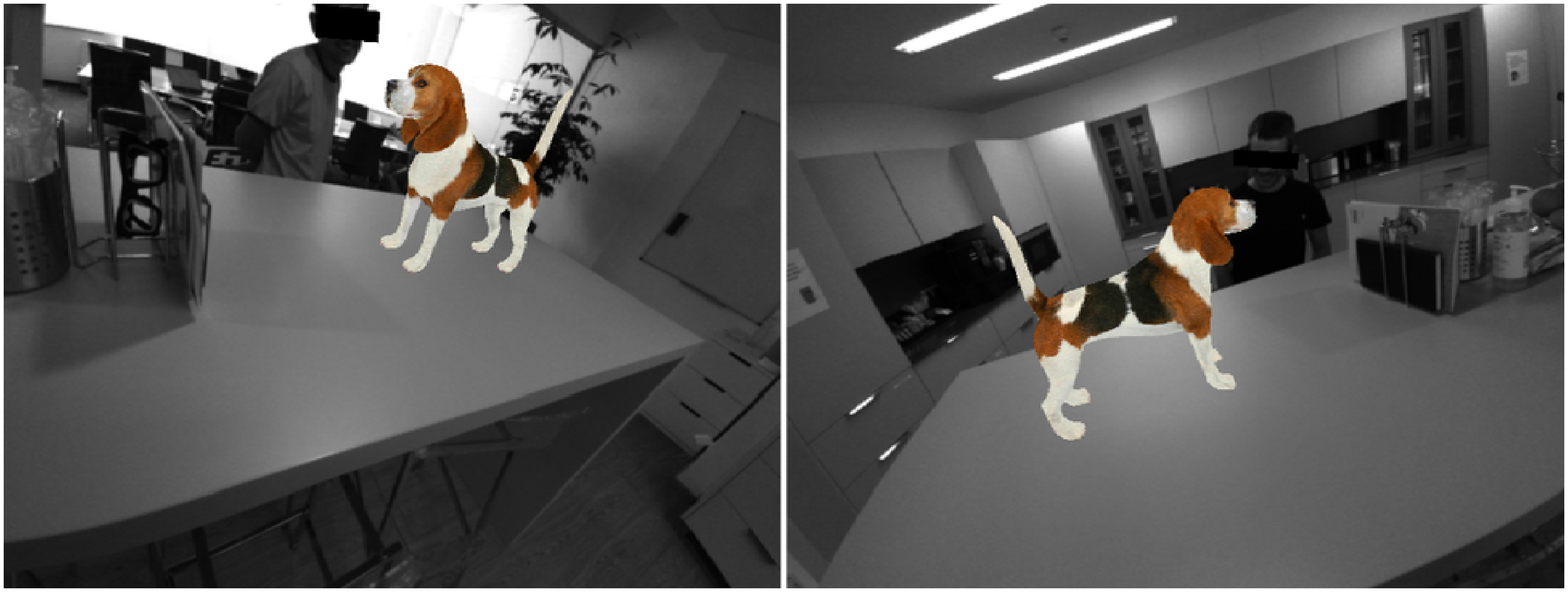} \\
		
	\end{tabular}
	\caption{\MM{kitchen} sequence. Top two rows: the picture to the left shows accuracy achieved after \MM{ransac} on minimal solver, using tracklets of the blue point. Glasses are censored for copyright reasons. The right image shows improved accuracy after the refinement stage as the pyramid aligns better to the marker. Bottom row: examples how a virtual 3\MM{d} object placed into the scene would be observed simultaneously by both users in their see-through displays.}
	\label{fig:kitchen}
\end{figure}

\begin{figure}
	\frame{\includegraphics[width = 0.5\linewidth]{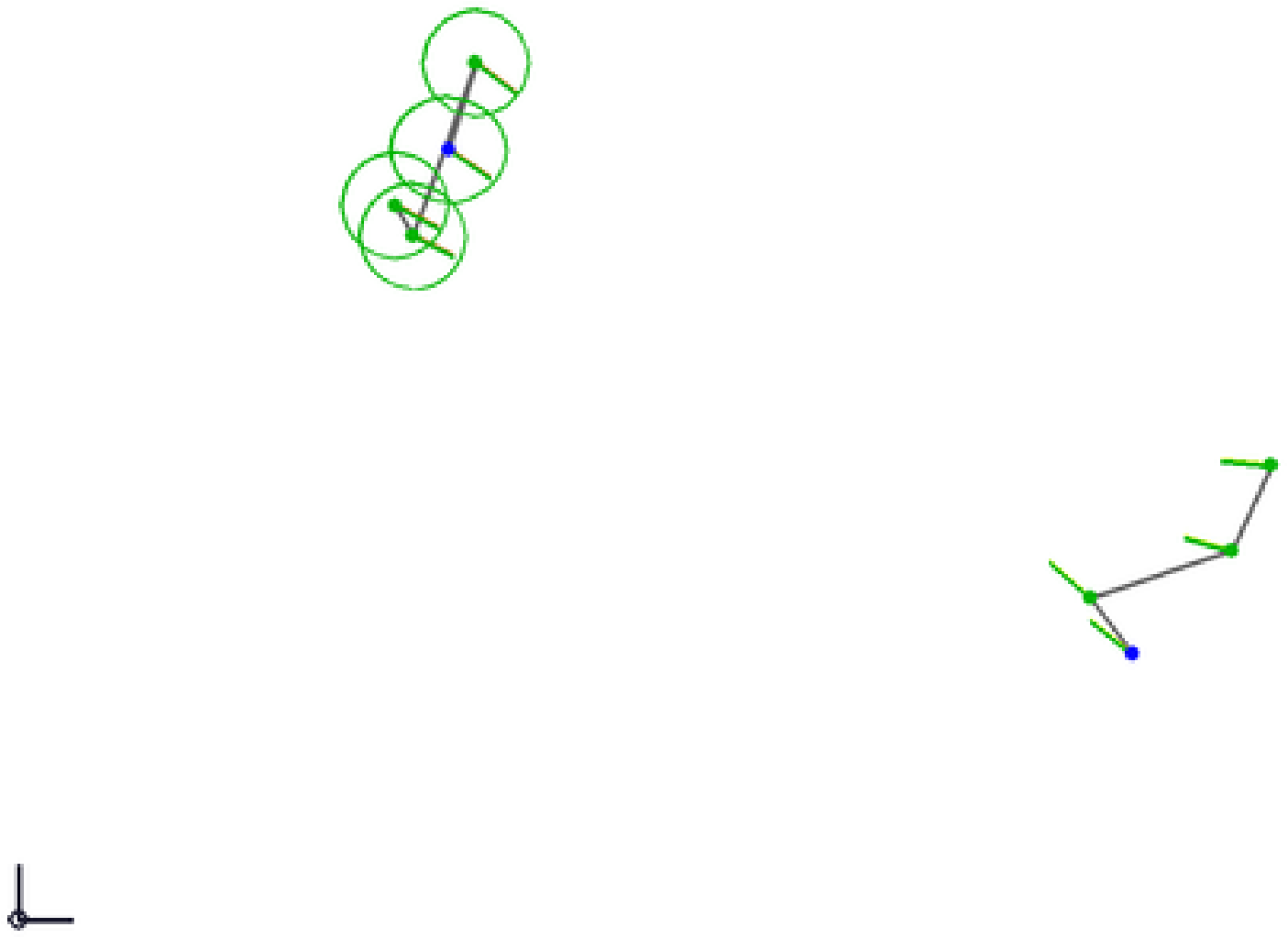}}
	\frame{\includegraphics[width = 0.5\linewidth]{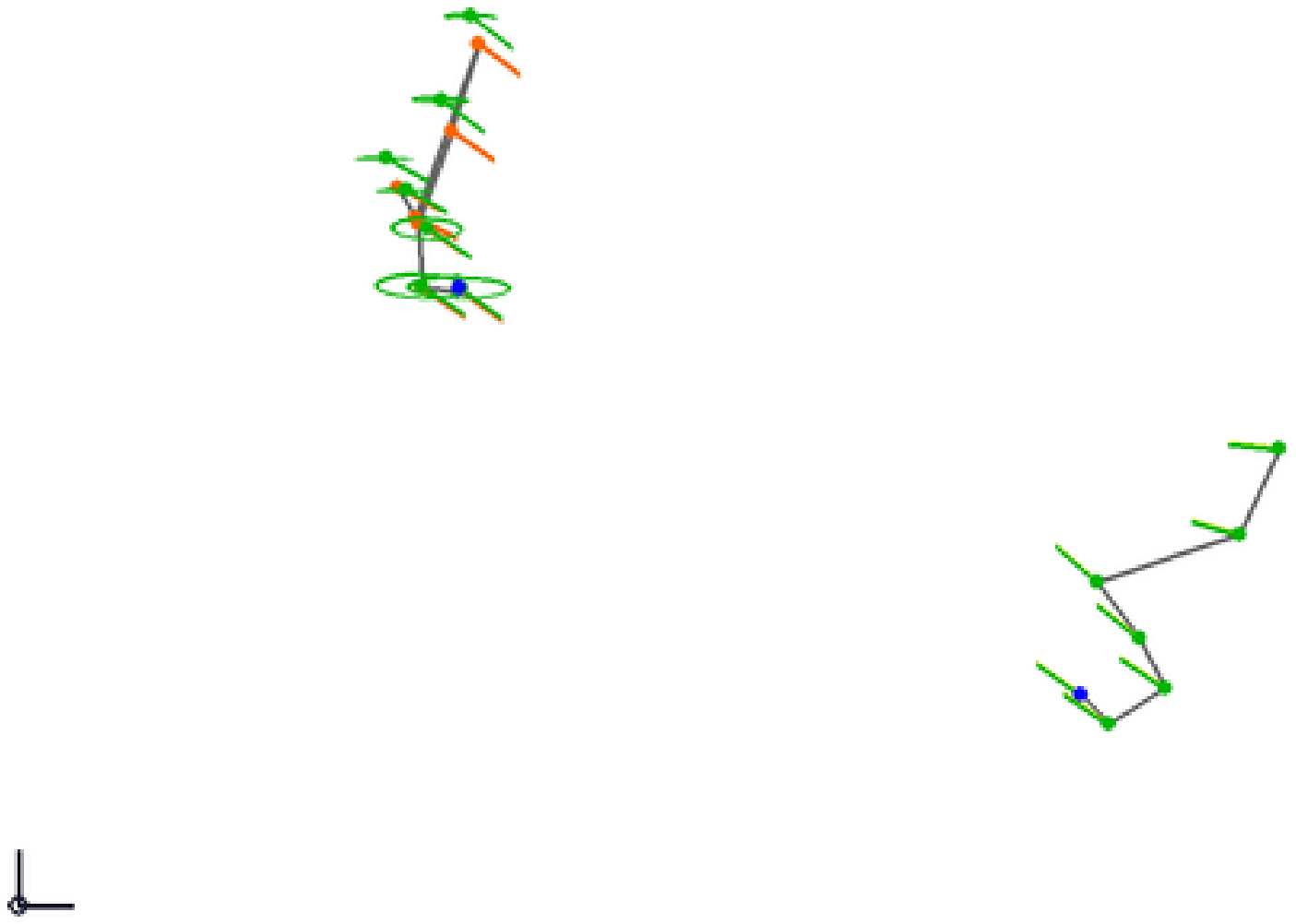}}\\[.5ex]
	\frame{\includegraphics[width = 0.5\linewidth]{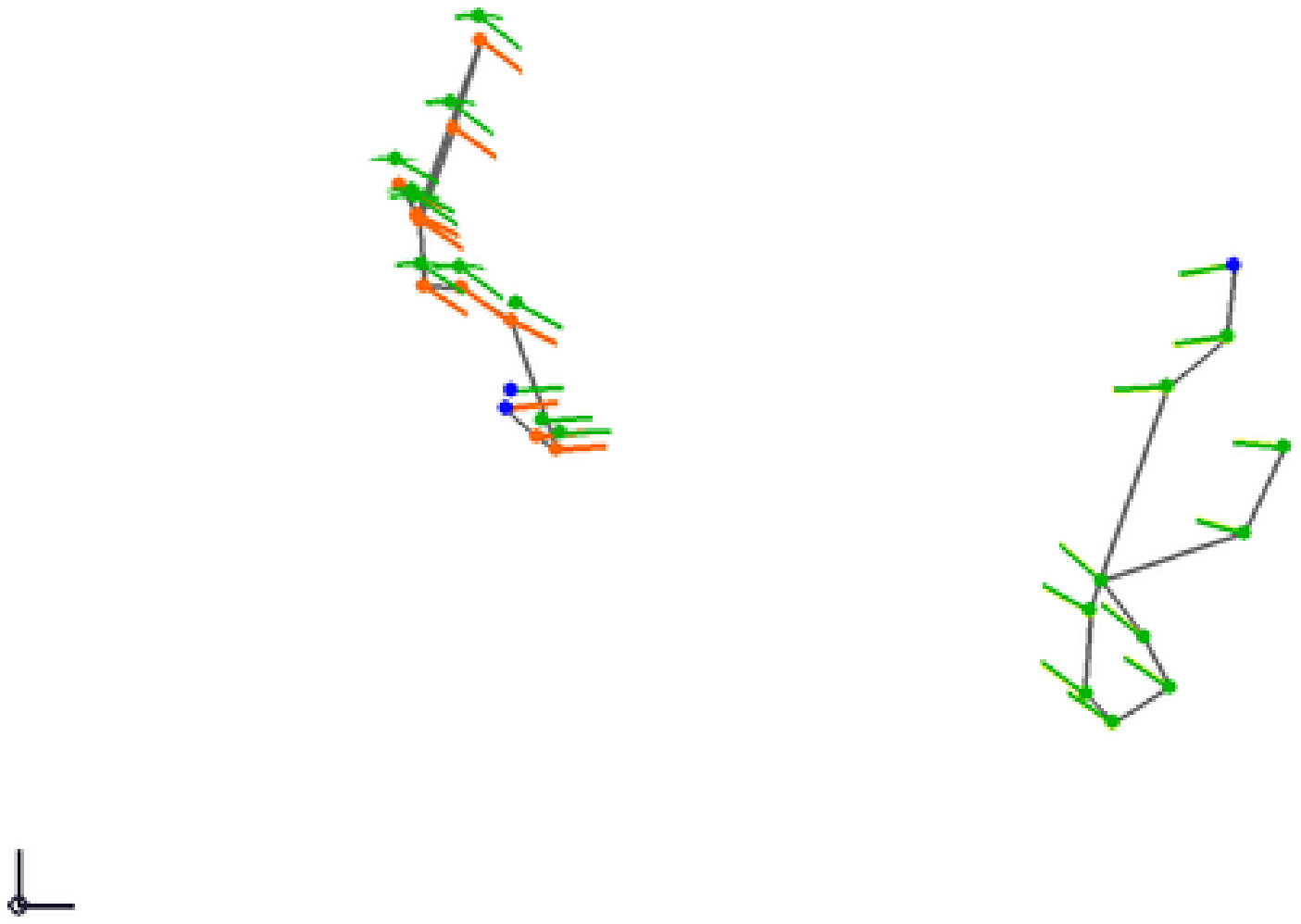}}
	\frame{\includegraphics[width = 0.5\linewidth]{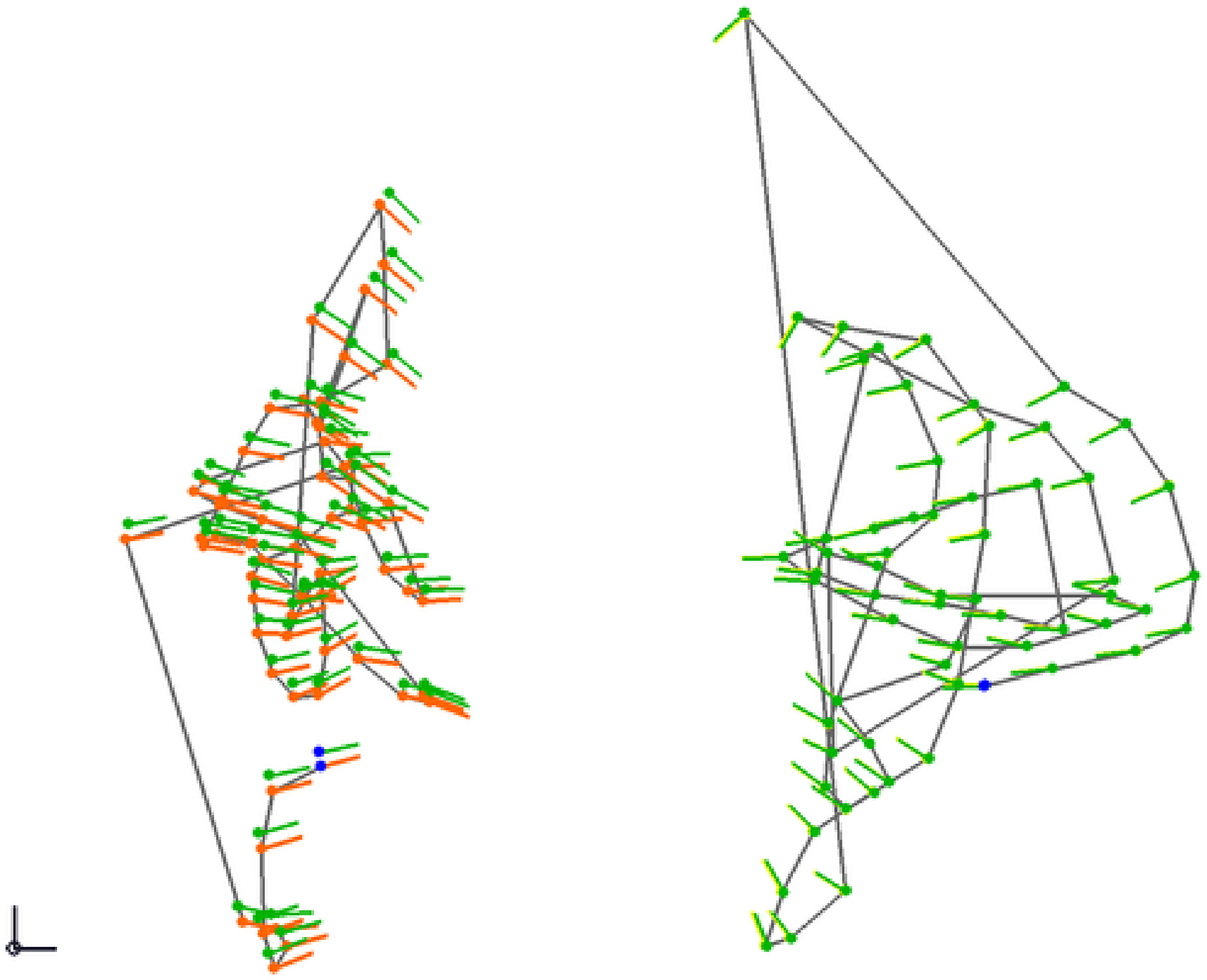}}
	\caption{\MM{kitchen} sequence, progress of GBP as new frames are added. The images show a top view on the two cameras with their looking direction, the left one of user $\C{B}$, the right of user $\C{A}$. In red are initial values, in green the refined ones. Note large covariance ellipses, \ie high uncertainty at the beginning of the sequence. The length of the coordinate axes is $0.1$m.}
	\label{fig:kitchen_gbp}
\end{figure}
We evaluate the method on Snap Spectacles glasses. They possess a rolling shutter camera and \imu{}. They are time synchronized and deliver $6$\MM{d} relative gravity aligned ego-poses from a VIO system. The camera of the glasses is calibrated internally and externally \wrt to the \imu{} sensor. In order to better visualize the achieved accuracy of the proposed solver, we augment virtual content into the images of the same camera which is used for $6$\MM{d} ego-pose estimation. In a real application on smart AR glasses, users would see the content in transparent see-through displays instead.

We track a point on the glasses with known $3$\MM{d} model which is the case of known lever arm with hard constraint [\MM{c}\oldstylenums{3}] from \Sec{sec:leverarm_constraint} and we utilize the bidirectional solver. To track the glasses, we use the technique described in \Sec{sec:glasses_tracing}. The cameras have VGA resolution with such a field of view which makes the face detection at the distance of $2$ meters between the users challenging. Width of the face can go down to about $50$ pixels which results to higher uncertainty than, \eg, one is used in multi-view reconstruction or SfM in computer vision domain. Detecting and tracking a point on the face was empirically pre-set to standard deviation of $1$ pixel. As an important result, we confirm that the presented method can handle this relatively high inaccuracy and still delivers sufficiently accurate relative pose for seamless augmentations. We experimentally confirmed that the super VGA resolution brings noticeable improvements as face and glasses detections become more accurate. The use of either resolution is a trade-off between accuracy, processing time and energy consumption. Choice of resolution depends on the specific use case and availability of the video stream on a specific device. We present only the result on  VGA resolution to demonstrate the worse case scenario. 

\begin{figure}
	\begin{center}
		\begin{tabular}{c@{\hspace{1mm}}c}
			\includegraphics[width = 0.49\linewidth]{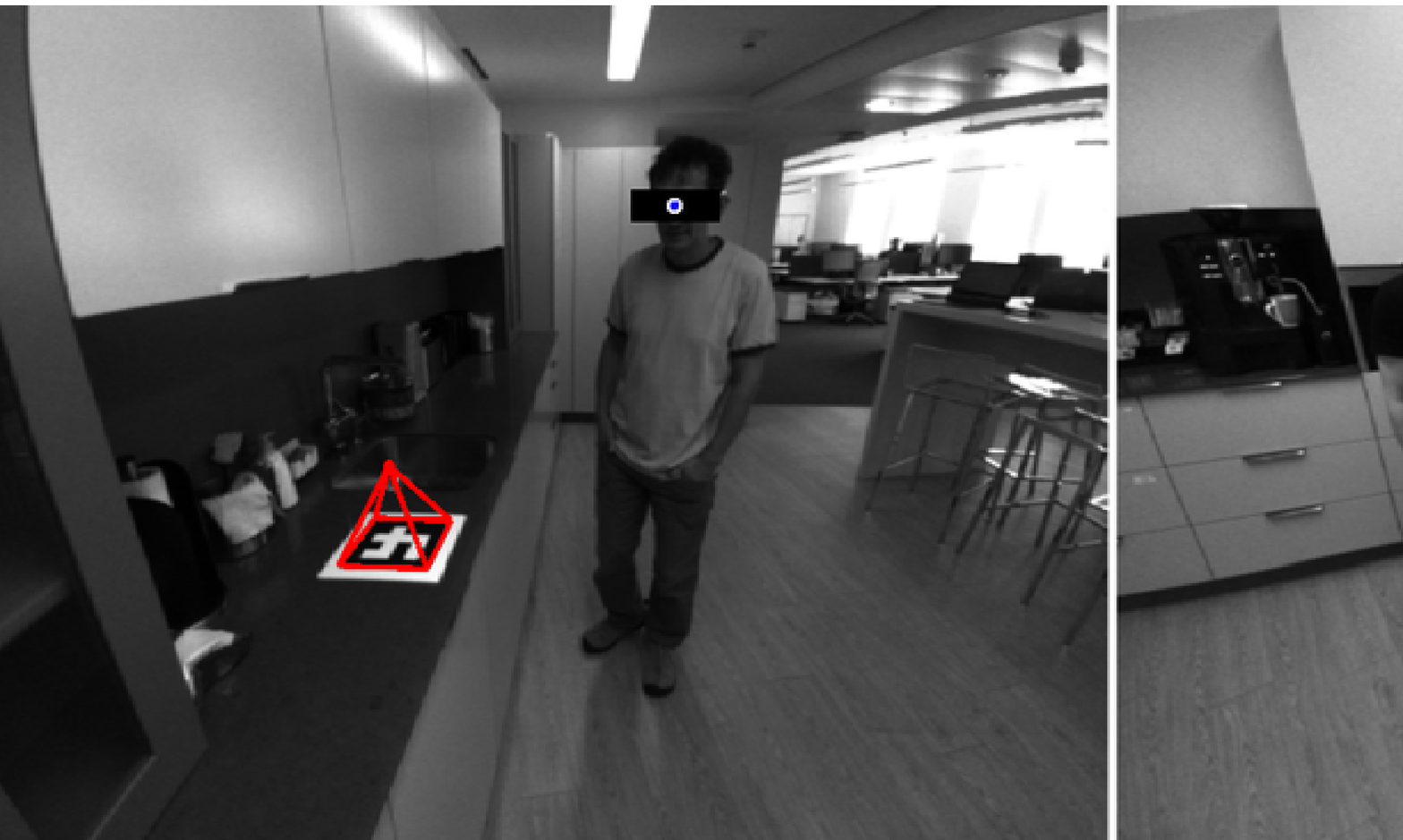} &
			\includegraphics[width = 0.49\linewidth]{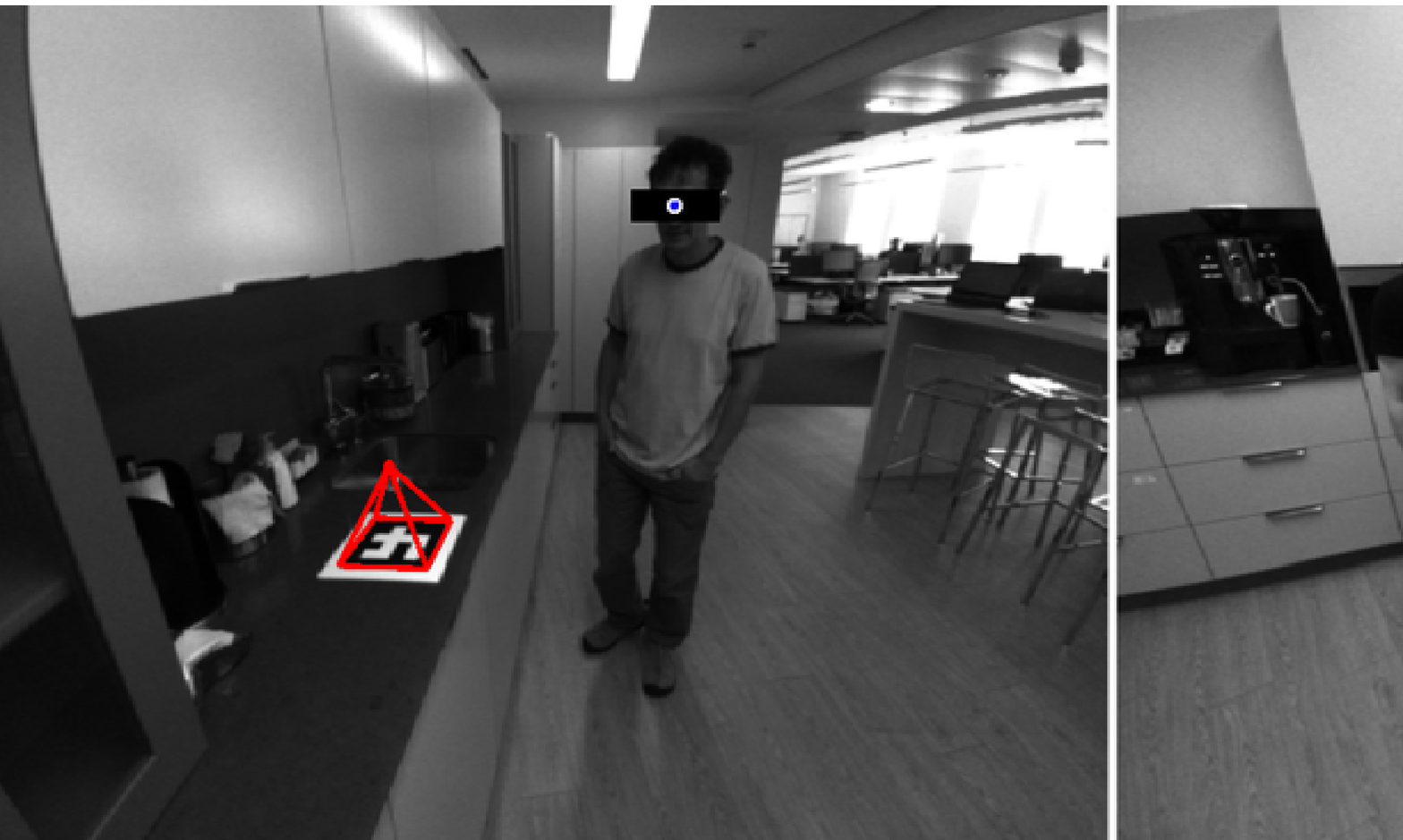} \\
			\includegraphics[width = 0.49\linewidth]{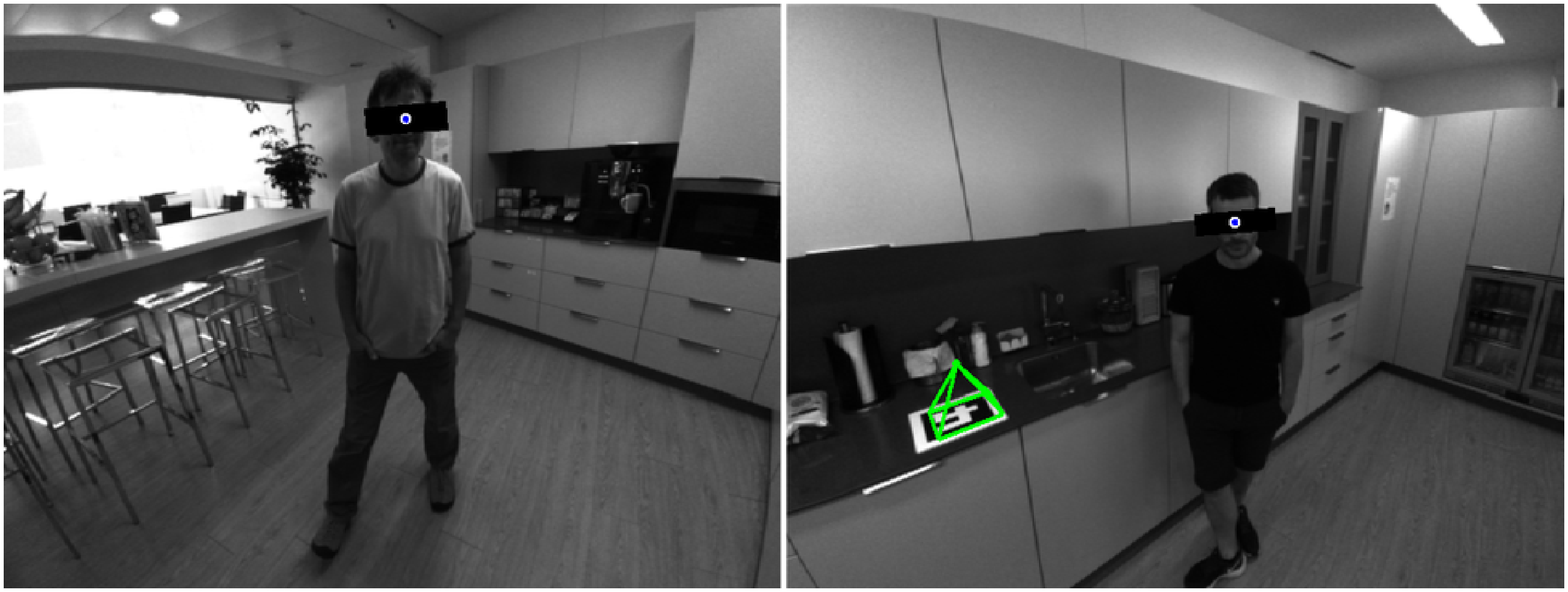} &
			\includegraphics[width = 0.49\linewidth]{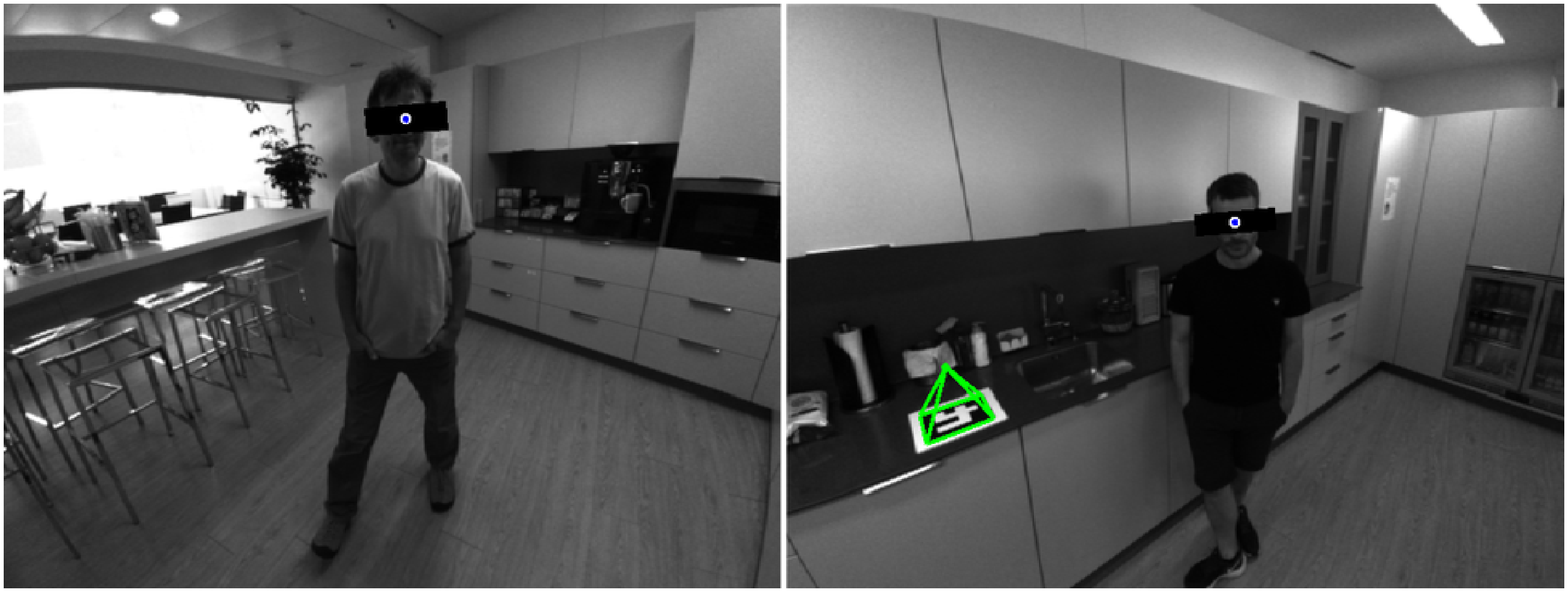} \\
			minimal solver & refinement \\[2ex]
		\end{tabular}
		\frame{\includegraphics[width = 0.7\linewidth]{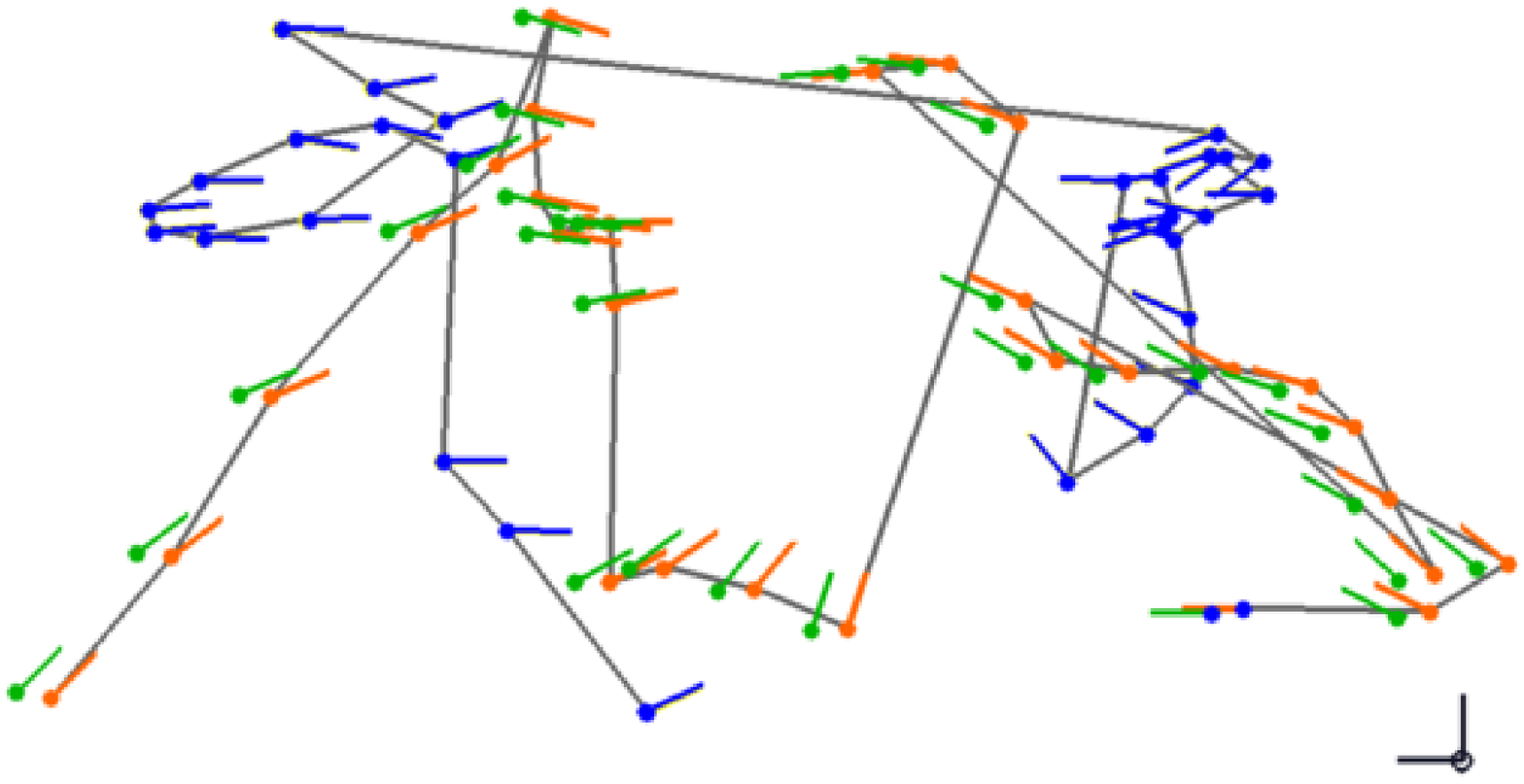}} 
	\end{center}\vspace*{-0.5cm}
	\caption{\MM{cupboard} sequence. In the bottom figure, user $\C{A}$ is plotted in blue to visually disentangle the two users.}
	\label{fig:cupboard}
\end{figure}

To better evaluate accuracy of the solver, we placed an AprilTag marker \cite{Olson-ICRA2011} into the scene at the beginning of the sequences. From the marker detections of the device of user $\C{A}$, we extract image coordinates of the four marker vertices and use them for their  multi-view triangulation given the poses. We constrain the triangulation such that the 3\MM{d} points lie on the plane which is perpendicular to the gravity vector. These four 3\MM{d} points are then used as basement for a virtual pyramid which can be then projected with known 6\MM{d} VIO ego-poses into the image views, see the red pyramid in the left image of \eg \Fig{fig:kitchen}. The inaccuracy of aligning the basement of the red pyramid \wrt the marker is mainly influenced by the little drift in VIO poses. Note that the pyramid lives in the local coordinate system of user $\C{A}$. The estimated relative transformation is then used to project the pyramid into images of user $\C{B}$, shown as a green pyramid in \Fig{fig:kitchen}. Assuming perfect VIO ego-poses, the more misaligned the basement of the projected green pyramid \wrt the real marker is, the more erroneous the estimated relative pose is. 

We present qualitative  as well as quantitative results on three sequences of roughly $1$ minute in length. We first run the minimal closed form solver in the iterative \MM{ransac} framework in order to cope with outliers. The estimated set of inliers is used to build and to solve more robust  overconstrained system. This estimate is fed to initialize the GBP refiner. Both, the minimal solver and the refinement are visualized in the corresponding figures for each sequence.

The first sequence is the \MM{kitchen} sequence. Two users look at the desk between them, while moving left and right by $1.5$ meters. Their trajectory can be seen in \Fig{fig:kitchen_gbp}. The minimal solver successfully estimates relative transformation, which is further improved by the refinement. It can be seen from the left bottom image in \Fig{fig:kitchen_gbp} that already after some seconds the refinement becomes certain which is seen by getting confidence ellipses smaller. Further frames do not contribute that much to the final estimate. After the relative transformation is estimated, a virtual dog is placed into the scene instead of the pyramid for a nicer user experience, see \Fig{fig:kitchen} bottom. For example, one user can point to a dog's body part while the other one can follow it.  

\begin{figure}
	\begin{center}
		\begin{tabular}{c@{\hspace{1mm}}c}
			\includegraphics[width = 0.49\linewidth]{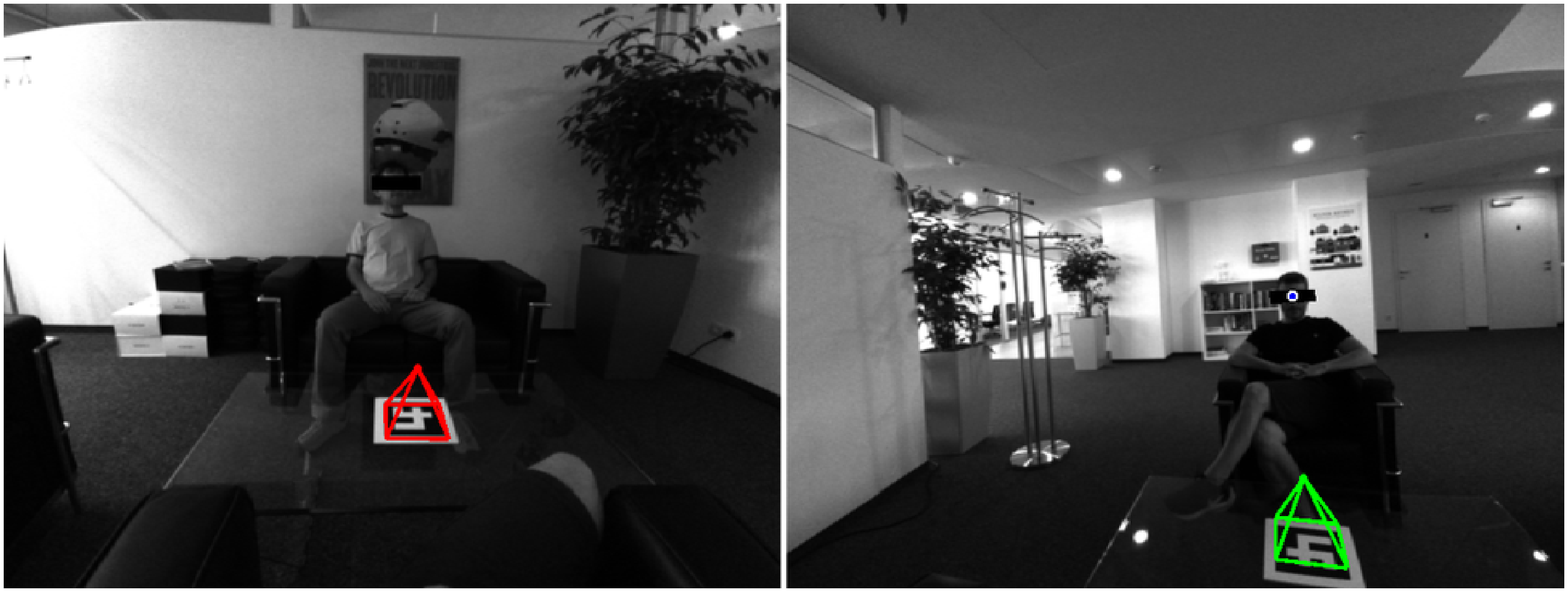} &
			\includegraphics[width = 0.49\linewidth]{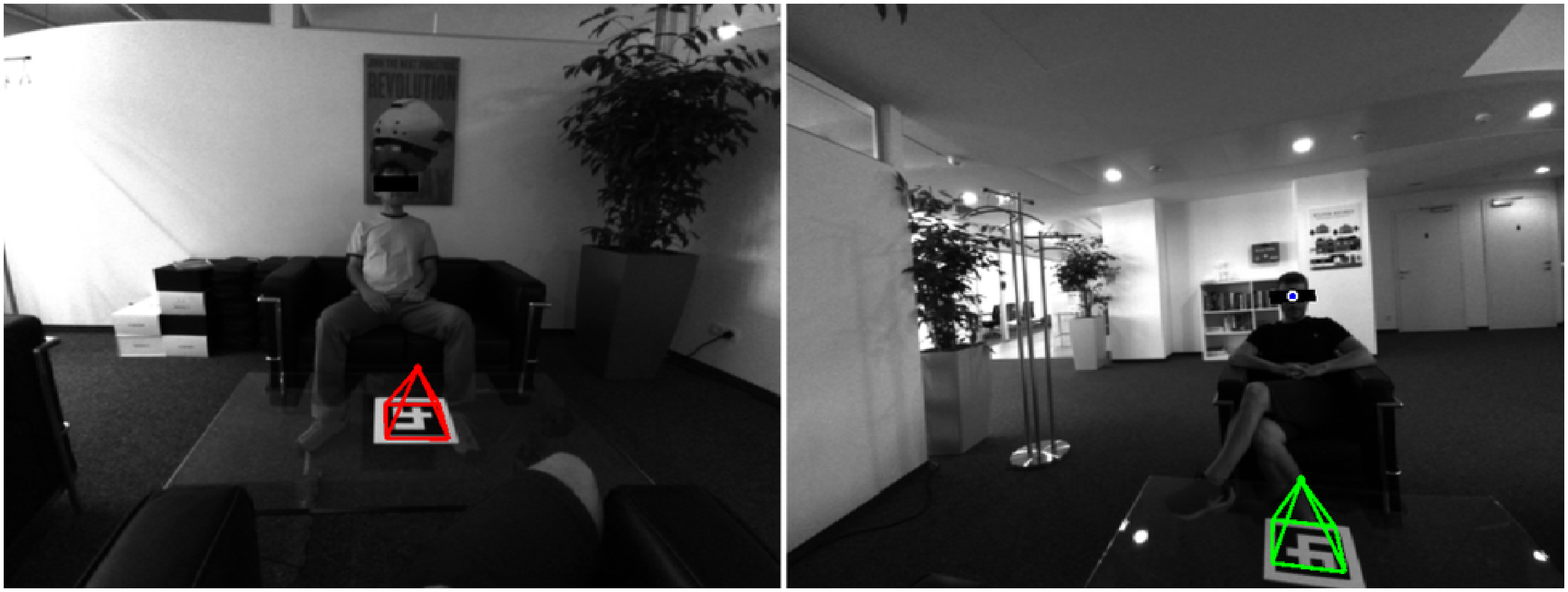} \\
			minimal solver & refinement \\[2ex]
			\includegraphics[width = 0.49\linewidth]{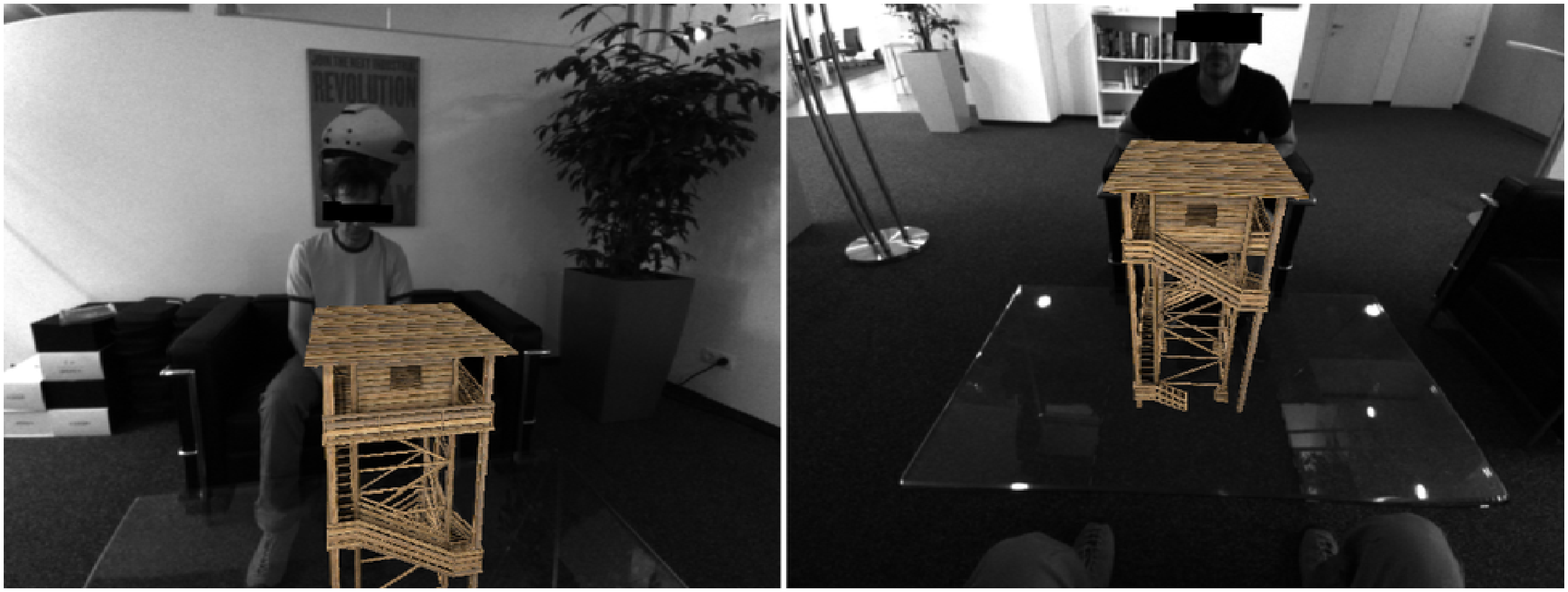} &
			\includegraphics[width = 0.49\linewidth]{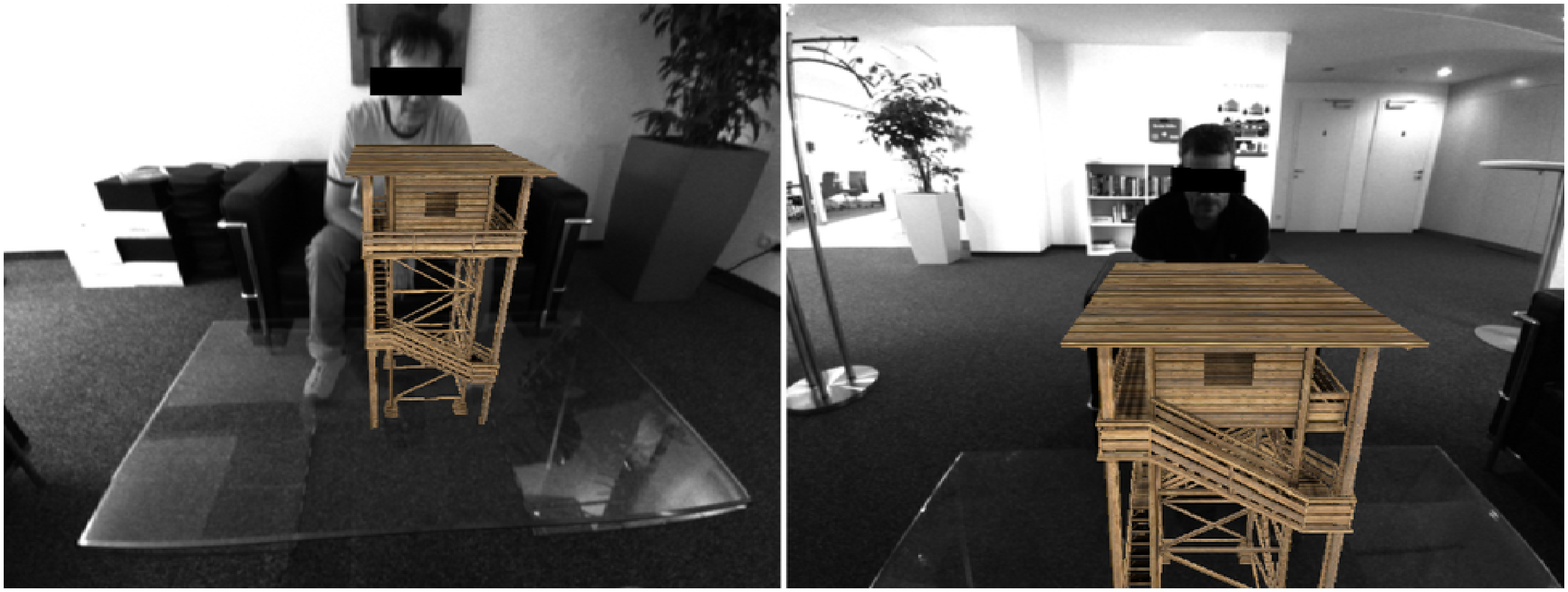} \\
		\end{tabular}
		\frame{\includegraphics[width = 0.7\linewidth]{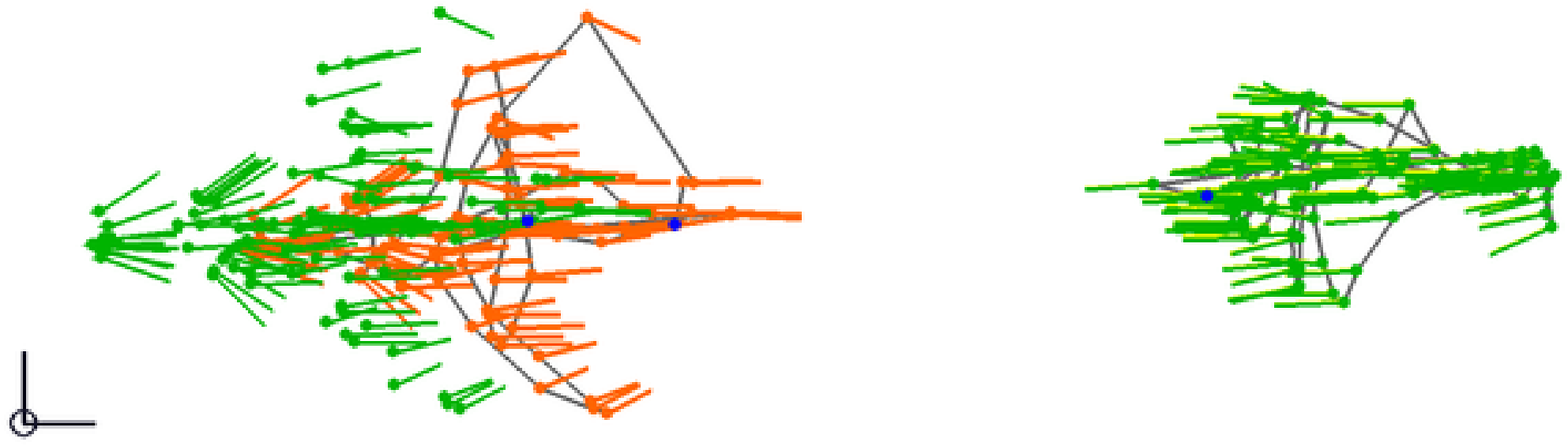}} 
	\end{center}\vspace*{-0.5cm}
	\caption{\MM{couch} sequence.}
	\label{fig:couch}
\end{figure}

\begin{table}
	\begin{center}
		\begin{tabular}{|l|c|c|c|} 
			\hline
			solver \textbackslash \ sequence & \MM{kitchen} & \MM{cupboard} & \MM{couch} \\\hline
			min solver error [pixel] & 9.8 & 5.2 & 15.1 \\
			refinement error [pixel] & 2.7 & 1.1  & 7.4 \\
			\hline
		\end{tabular}
	\end{center}
	\caption{Quantitative comparison on the presented sequences. For each sequence, roughly $200$-$250$ images with successful marker detections were used to calculate the error.}
	\label{tab:quantitative_comparison}
\end{table}

The second sequence is the \MM{cupboard} sequence. The users look at the cupboard from similar viewpoints while walking around within the range of $2$ meters while looking sporadically at each other, see \Fig{fig:cupboard} for results. The initial solution is already good, while the refinement tunes it further to get the estimate very accurate.  

The third sequence is the \MM{couch} sequence. The users sit on couches with a coffee table in between and slightly move their heads within the range of $0.5$ meter, see \Fig{fig:couch}. Again, after the relative transformation is estimated, a virtual watchtower is placed into the scene instead of the pyramid for a more pleasing user experience. We see this sequence performing the worst although satisfactory for the targeted use case. The refinement in this case cannot fully recover. The reason is that this type of motion brings small baseline for involved geometric constructions and let the noise on tracklets dominate. It is close to the critical configuration when the users would not translate their heads but only rotate. The inaccuracy is unobservable for the user when looking at the virtual content, the watchtower in this case.

For quantitative comparison, we evaluated median error as the image distance in camera $\C{B}$ between projected vertices of the pyramid basement via the estimated relative transformation and the actual detection of the marker corner points. The statistics can be seen in \Tab{tab:quantitative_comparison}.  We first reconstruct the marker in $3$\MM{d} for user $\C{A}$ with his given VIO ego-poses. If VIO slightly drifts, the $3$\MM{d} pyramid might be reconstructed slightly inaccurately which propagates further. Even if the relative pose would be perfectly estimated, and there was no VIO drift for user $\C{B}$, the evaluation metric would still report a misalignment. 

Overall, the minimal solver scores in reasonable bounds, considering inaccuracy of the glasses detector. The refinement noticeably improves on it further. Note that this evaluation metric includes multiple errors in the whole pipeline and gives clear intuition where the final accuracy lies for practical scenarios.
 
\paragraph{Discussion}
Note that the \MM{kitchen} and \MM{couch} sequences represent scenarios where standard visual localization methods within the point cloud are prone to fail. Reason is that the glasses cameras observe the scene points from very different oblique views and that overlapping seen area is very small. Typical image descriptors are not invariant to such view angle differences which makes the standard $2$\MM{d}-$3$\MM{d} matching very hard if not impossible. On the other hand, the proposed methodology successfully deals with such cases and provides a  lightweight solution for seamless collaborative AR. As a result, the users do not need to move and orient themselves in a way such that the standard visual localization works. Instead, they would freely move and look at each other like in a typical social interaction scenario.

\section{Conclusions}
We presented a novel practical solution for the ego-motion alignment problem. We introduced necessary conditions and math to derive a closed form solver for the alignment problem just from tracking faces and worn glasses of AR users. As one of the novelties we show that tracklets can serve as reliable anchors to estimate the unknown relative alignment of local coordinate systems with sufficient accuracy for practical use. The proposed methodology offers a lightweight solution which is well suited for today's smart AR glasses. It complements traditional visual localization methods which require much more computational and memory resources. Specifically, the proposed method aims at handling situations when users are close to and facing each other, and collaborating on a shared virtual object. The experimental results on the synthetic as well as on the real data show its high practical potential.

\bibliographystyle{abbrv} 
\bibliography{egoMotionAlignment}

\end{document}